%% file: humanoids2018.tex
\documentclass[letterpaper, 10 pt, conference]{ieeeconf}  

\IEEEoverridecommandlockouts                              

\usepackage{graphicx}
\usepackage{subcaption} 

\usepackage{amssymb}
\usepackage{amsmath}
\usepackage{commath}
\usepackage{mathtools}

\usepackage{hyperref}
\usepackage{import}
\usepackage{paralist}
\usepackage{gensymb}
\usepackage{dsfont}
\usepackage{siunitx}
\usepackage[bottom]{footmisc}
\usepackage{framed}
\usepackage{balance} 
\usepackage{caption}
\usepackage[skins]{tcolorbox}

\newtcolorbox{myframe}[2][]{%
  enhanced,colback=white,colframe=black,coltitle=black,
  sharp corners,boxrule=0.4pt,left=0pt,right=0pt,top=0pt,bottom=0pt,
  fonttitle=\itshape,
  attach boxed title to top left={yshift=-0.3\baselineskip-0.4pt,xshift=2mm},
  boxed title style={tile,size=minimal,left=0.5mm,right=0.5mm,
  colback=white,before upper=\strut},
  title=#2,#1
}

\newcommand*\diff{\mathop{}\!\mathrm{d}}

\DeclareCaptionFont{mysize}{\fontsize{8}{9.6}\selectfont}
\usepackage{graphics} 
\usepackage{epsfig} 
\usepackage{todonotes}
\usepackage[normalem]{ulem}
\DeclareMathOperator*{\argmin}{arg\,min}

\graphicspath{{figures/}}

\title{\LARGE \bf
 A Benchmarking of DCM Based Architectures \\ for Position and Velocity Controlled  Walking of Humanoid Robots
}

\author{Giulio Romualdi, Stefano Dafarra, Yue Hu, Daniele Pucci$^{1}$
  \thanks{$^{1}$ Romualdi, Dafarra, Hu and Pucci are with the Fondazione Istituto Italiano
    di Tecnologia, 16163 Genova, Italy (e-mail: name.surname@iit.it).}
}

\begin{document}

\maketitle
\thispagestyle{empty}
\pagestyle{empty}

\begin{abstract}
This paper contributes towards the development and comparison of Divergent-Component-of-Motion (DCM) based control architectures for humanoid robot locomotion. More precisely, we present and compare several DCM based implementations of a  three layer control architecture. From top to bottom, these three layers are here called: \emph{trajectory optimization}, \emph{simplified model control}, and \emph{whole-body QP control}. 
All layers use the DCM concept to generate references for the layer below. For the \emph{simplified model control} layer, we present and compare both instantaneous and Receding Horizon Control controllers. For the  \emph{whole-body QP control} layer, we present and compare controllers for  position and velocity controlled robots. Experimental results are carried out on the one-meter-tall iCub humanoid robot. We show which implementation of the above control architecture allows the robot to achieve a walking velocity of~0.41 meters per second.

\end{abstract}

\input{tex/introduction.tex}
\input{tex/background.tex}
\input{tex/architecture.tex}
\input{tex/results.tex}

\input{tex/conclusion.tex}
\balance








\bibliography{bibliography}
\bibliographystyle{IEEEtran}

\end{document}

%% file: tex/introduction.tex
\section{INTRODUCTION}


Bipedal locomotion of humanoid robots remains an open problem despite decades of research in the subject. The complexity of the robot dynamics, the unpredictability of its surrounding environment, and the low efficiency of the robot actuation system are only a few problems that complexify the achievement of robust robot locomotion. In the large variety of robot controllers for bipedal locomotion, the Divergent-Component-of-Motion (DCM) is an ubiquitous concept used for generating walking patterns. This paper presents and compares different DCM based control architectures for humanoid robot locomotion.


\par

During the DARPA Robotics Challenge, a common approach for humanoid robot control was that of defining an hierarchical  architecture composed of several layers~\cite{feng2015optimization}. Each layer generates references for the layer below by processing inputs from the robot, the environment, and the outputs of the layer before. From top to bottom, these layers are here called: \emph{trajectory optimization}, \emph{simplified model control}, and \emph{whole-body quadratic programming (QP) control}.

The \emph{trajectory optimization} layer often generates  desired foothold locations by means of optimization techniques. To do so, both kinematic and dynamical robot models can be used~\cite{dai2014whole,herzog2015trajectory}. When solving the optimization problem associated with the \emph{trajectory optimization} layer, computational time may be a concern especially when the robot surrounding environment is not structured. There are cases, however, where simplifying assumptions on the robot environment can be made, thus reducing the associated computational time. For instance, flat terrain allows one the view the robot as a simple unicycle \cite{PascalHandbook,flavigne2010reactive}, which enables quick solutions to the optimization problem for the walking pattern generation~\cite{Dafarra2018}. 
\par
\begin{figure}[t]
  \centering
      \begin{subfigure}[b]{0.29\columnwidth}
        \centering
        \includegraphics[width=\textwidth]{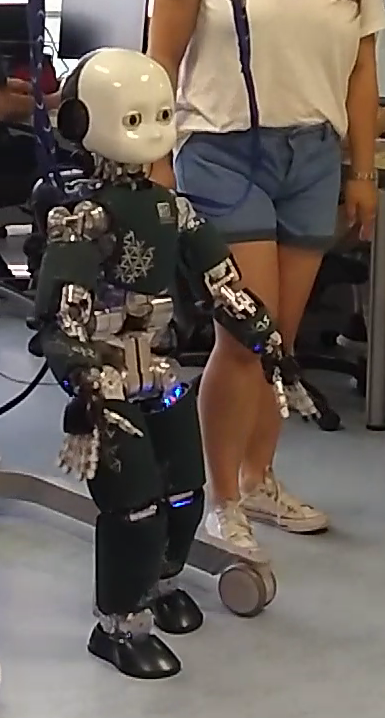}
    \end{subfigure}
    \begin{subfigure}[b]{0.29\columnwidth}
        \centering
        \includegraphics[width=\textwidth]{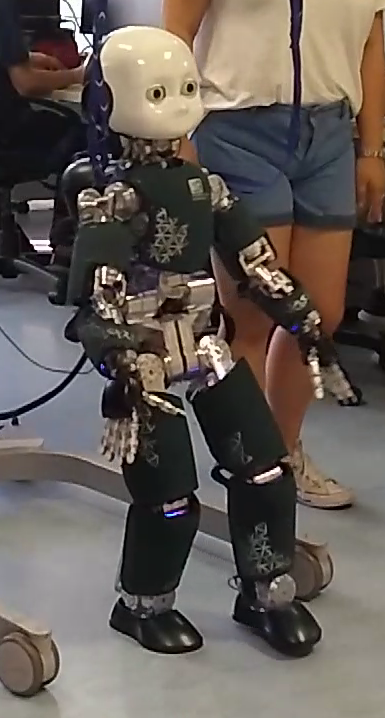}
    \end{subfigure}
    \begin{subfigure}[b]{0.29\columnwidth}
        \centering
        \includegraphics[width=\textwidth]{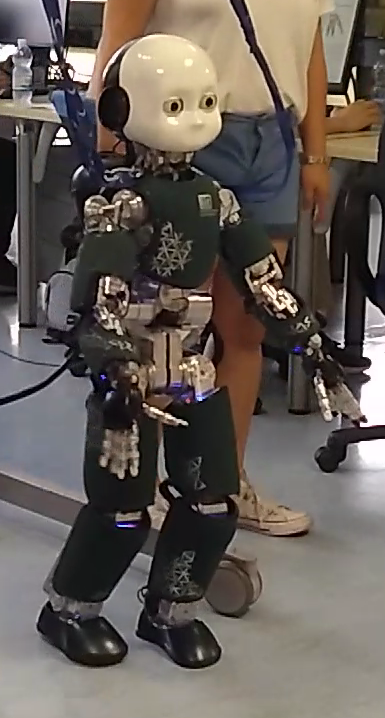}
    \end{subfigure}
  \caption{iCub walks with the presented controller architecture. \label{fig:icub}}
      \vskip-0.5cm
\end{figure}
The \emph{simplified model control} layer
is in charge of finding feasible center-of-mass (CoM) trajectories and is often based on simplified dynamical models, such as the Linear Inverted Pendulum Model (LIPM) \cite{Kajita2001} and the Capture Point (CP) \cite{Pratt2006}. These models have become very popular after the introduction of the Zero Moment Point (ZMP)  as a stability criterion~\cite{Vukobratovic1969a}. 
To obtain feasible CoM trajectories, the \emph{simplified model control} layer often combines the LIPM with Model Predictive Control (MPC) techniques, also known as the Receding Horizon Control (RHC) \cite{Kajita2003,diedam2008online}. 
Another model that is often exploited in the \emph{simplified model control} layer is the Divergent Component of Motion (DCM) \cite{Englsberger2015}. 
The DCM can be viewed as the extension of the capture point (CP) to the three dimensional case, however always under the assumption of a constant Virtual Repellent Point (VRP) to Enhanced Centroidal Moment Pivot point (eCMP) height difference \cite{Englsberger2015}.
Attempts at loosening this latter assumption and extending the DCM to more complex models have also been presented \cite{Hopkins2015}. 

The \emph{whole-body QP control} layer generates robot  positions, velocity or torques depending on the available control modes of the underlying robot. These outputs aim at stabilizing the references generated by the layers before. It uses whole-body kinematic or dynamical models, and very often instantaneous optimization techniques, namely, no MPC methods are here employed. Furthermore, the associated optimisation problem is often framed as an hierarchical stack-of-tasks, with strict or weighted hierarchies \cite{Stephens2010,Nava2016}.

This paper presents and compares several DCM based implementations of the above layered control architecture. 
In particular, the \emph{trajectory optimization} layer is kept fixed with a unicycle based  planner that generates desired DCM and foot  trajectories. The \emph{simplified model control} layer, instead, implements two controllers for the tracking of the DCM: an instantaneous and an MPC controller. 
In the same layer, we also present a controller that ensures the tracking of the CoM and the ZMP, which exploits 6-axes Force Torque sensors (F/T).
Finally, the \emph{whole-body QP control} ensures the tracking of the desired CoM  and feet trajectories. It achieves this by presenting velocity and inverse kinematic based controllers. 
The several combinations of the control architecture are tested on the iCub humanoid robot \cite{Nataleeaaq1026} .
\par
The paper is organized as follows. Sec.~\ref{sec:BACKGROUND} introduces notation, the humanoid robot model, and some simplified models used for locomotion.
Sec.~\ref{sec:ARCHITECTURE} describes each layer of the control
architecture, namely the trajectory optimization, the simplified model control and and the whole-body QP control layer. Sec.~\ref{sec:EXPERIMENTAL_RESULTS} presents the experimental validation of the
proposed approach, and shows an explanatory table comparing the different control approaches. Finally, Sec.~\ref{sec:CONCLUSION_AND_FUTURE_WORK} concludes the paper.

%% file: tex/background.tex
\section{BACKGROUND}
\label{sec:BACKGROUND}
\subsection{Notation}
\begin{itemize}
\item $I_n$ and $0_n$ are used to denote respectively the $n\times n$ identity and zero matrices;
\item $\mathcal{I}$ denotes an inertial frame;
\item $e_i \in \mathbb{R}^n$ is the canonical vector, consisting of all zeros
  but the $i$-th component that is equal to one;
\item given two frames, $\mathcal{A}$ and $\mathcal{B}$, $\prescript{\mathcal{A}}{}{R}_B \in SO(3)$ represents the rotation matrix between the frames, i.e.
  given two vectors $\prescript{\mathcal{A}}{}{p}, \prescript{\mathcal{B}}{}{p} \in \mathbb{R}^3$ respectively expressed in $\mathcal{A}$ and $\mathcal{B}$,
  the rotation matrix $\prescript{\mathcal{A}}{}{R}_\mathcal{B}$ is such that $\prescript{\mathcal{A}}{}{p} = \prescript{\mathcal{A}}{}{R}_\mathcal{B} \prescript{\mathcal{B}}{}{p}$;
\item given a skew-symmetric matrix $W\in \mathfrak{so}(3)$ the \emph{vee operator} is $.^\vee : \mathfrak{so}(3) \to \mathbb{R}^3$;
\item $\prescript{\mathcal{A}}{}{\omega}_\mathcal{B} \in \mathbb{R}^3$ denotes the angular velocity
  between the frame $\mathcal{B}$ and the frame $\mathcal{A}$ expressed in the frame $\mathcal{A}$;
\item given a square matrix $A \in \mathbb{R}^{3 \times 3}$ the \emph{skew} operator is  $\text{sk} :\mathbb{R}^{3 \times 3} \to \mathfrak{so}(3)$, $\text{sk}(A) := (A - A^\top)/2$;
\item the subscripts $\mathcal{T}$, $\mathcal{LF}$, $\mathcal{RF}$ and
$\mathcal{C}$ indicates the frames attached to the torso, left foot, right foot and CoM.
\end{itemize}
\subsection{Humanoid robot models}
A humanoid robot is an example of floating base multibody systems composed of $n + 1$ links connected by $n$ joints with
one degree of freedom.
Since none of the links of the robot has an a priori constant position and orientation with respect to a global frame $\mathcal{I}$,  its configuration can be determined by the position
and the orientation of the base frame $\mathcal{B}$ (e.g. the stance foot during the locomotion) and the joints values. Thus the configuration space is defined by
$\mathbb{Q} = SO(3) \times \mathbb{R}^3 \times \mathbb{R}^n$. An element of $\mathbb{Q}$
is then a triplet $q = (\prescript{\mathcal{I}}{}{p}_\mathcal{B}, \prescript{\mathcal{I}}{}{R}_\mathcal{B}, s)$,
where $(\prescript{\mathcal{I}}{}{p}_\mathcal{B}, \prescript{\mathcal{I}}{}{R}_\mathcal{B}) \in (SO(3) \times \mathbb{R}^3)$ is used to represent the position and the orientation of the base frame, $\mathcal{B}$, expressed
with respect to the inertial frame $\mathcal{I}$; and $s \in \mathbb{R}^n$ represents the joint
angles. Furthermore, according to the group theory, $\mathbb{Q}$ is a Lie group. Indeed given two
elements $a = (p_a, R_a, s_a) \in \mathbb{Q}$ and $b = (p_b, R_b, s_b)\in \mathbb{Q}$ the group
multiplication $a \cdot b$ is defined
as $a \cdot b = (p_a + p_b, R_a R_b, s_a + s_b) \in \mathbb{Q}$, while the inverse of $a$ is given by:
$a^{-1} = (-p_a, R_a^\top, -s_a) \in \mathbb{Q}$.
Since $\mathbb{Q}$ is a Lie group, the velocity of the multibody system is represented by the Lie algebra $\mathbb{V}$. An element of the Lie algebra $\mathbb{V}$ is a triplet
$ \nu = (\prescript{\mathcal{I}}{}{\dot{p}}_\mathcal{B}, \prescript{\mathcal{I}}{}{\omega}_\mathcal{B}, \dot{s})$
where $\prescript{\mathcal{I}}{}{\omega}_{\mathcal{B}}$ is the angular velocity of the base with respect the inertial frame whose coordinates are written in the inertial frame, i.e.
$\prescript{\mathcal{I}}{}{\dot{R}}_\mathcal{B} = \prescript{\mathcal{I}}{}{\hat{\omega}}_\mathcal{B} \prescript{\mathcal{I}}{}{R}_\mathcal{B}$.
\par
Given a link of the floating base system its position and orientation with respect to the inertial frame is uniquely identified by an homogeneous transformation, $\prescript{\mathcal{I}}{}{H}_\mathcal{A} \in SE(3)$,
between the inertial frame, $\mathcal{I}$, and the frame attached to that link, $\mathcal{A}$.
$\prescript{\mathcal{I}}{}{H}_\mathcal{A}$ is the classical homogeneous transformation containing the rotation matrix, $\prescript{\mathcal{I}}{}{R}_\mathcal{A}$,
and the vector connecting the origin of the inertial frame to origin of the frame $\mathcal{A}$
expressed in the inertial frame $\prescript{\mathcal{I}}{}{p}_\mathcal{A}$.

Given a link of the floating base system the velocity of a rigid body can be represented by $\prescript{\mathcal{I}}{}{v}_\mathcal{A} =
\begin{bmatrix}
  \prescript{\mathcal{I}}{}{\dot{p}}_\mathcal{A}^\top &
  \prescript{\mathcal{I}}{}{\omega}_{\mathcal{A}}^\top
\end{bmatrix}^\top$.
\par
The Jacobian $J_{\mathcal{A}}(q)$ is the map between the robot velocity and the linear and angular
velocities of the frame $\mathcal{A}$, i.e.
$\prescript{\mathcal{I}}{}{v}_\mathcal{A} = J_{\mathcal{A}} \nu$ where $J_{\mathcal{A}}$ can be split into
two parts, one related to the velocity of the base and the
other one multiplying the joints velocity, $J_\mathcal{A}(q) = \begin{bmatrix}   J^b_\mathcal{A} &   J^j_\mathcal{A} \end{bmatrix}$. 
\begin{figure*}[!t]
  \centering
\includegraphics[scale=0.89]{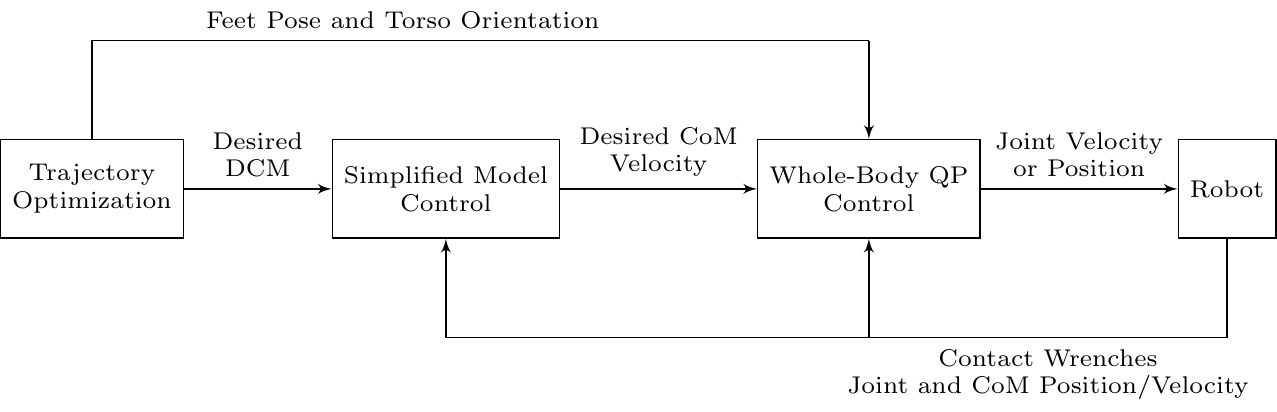}
  \caption{The control architecture is composed of three layers: the \emph{trajectory optimization}, the \emph{simplified model control}, and the \emph{whole-body QP control}.
  \label{fig:controller_architecture}}
      \vskip-0.5cm
\end{figure*}
\subsection{Simplified models}
\label{sub_sec:simplified_models}
For the purpose of this work, the motion of the humanoid robot is approximated by means of the well known
\emph{Linear inverted pendulum model} (LIPM) \cite{Kajita2001}.
By using the LIPM, 
in case of walking on a flat surface the CoM belongs to an horizontal plane with a constant height $z_0$.
The simplified CoM dynamics is given by \cite{Kajita2001}:
\begin{equation}
  \label{eq:3d-lipm}
  \ddot{x} = \omega^2(x - r^{zmp}),
\end{equation}
where $x \in \mathbb{R}^2$ is the vector containing the projection of the CoM on the walking surface, $r^{zmp}\in \mathbb{R}^2$ is the position
of the zero moment point (ZMP) and $\omega$ is the inverse of the pendulum time constant, i.e.
$\omega = \sqrt{g/z_0}$ where $g$ is the gravity constant.

Analogously, one can define the divergent component of motion
(DCM) as $\xi = x + \dot{x}/\omega$ \cite{Englsberger2015}. Clearly,  the DCM time derivative is given by:
\begin{equation}
  \label{eq:dcm_dynamics}
  \dot{\xi} = \omega(\xi - r^{zmp}).
\end{equation}
Using the DCM as state variable, the LIPM dynamics \eqref{eq:3d-lipm} can be decomposed into two
parts:
\begin{equation}
  \label{eq:simplified-model}
\begin{bmatrix}
  \dot{x}\\
  \dot{\xi}
\end{bmatrix} =
\begin{bmatrix}
  -\omega I_2 & \omega I_2\\
  0_2 & \omega I_2
\end{bmatrix}
\begin{bmatrix}
  x\\
  \xi
\end{bmatrix}
+
\begin{bmatrix}
  0_2\\
  \omega I_2
\end{bmatrix}
r^{zmp}.
\end{equation}
Performing the state space decomposition we can easily show that the CoM tends the DCM, while the
DCM dynamics has a strictly positive real part eigenvalue.

%% file: tex/architecture.tex
\section{ARCHITECTURE}
\label{sec:ARCHITECTURE}

This section summarizes the components of the control architecture presented in Fig. \ref{fig:controller_architecture}.
In particular, the control architecture is composed of three main layers.
The first layer is represented by the \emph{trajectory generator}, whose main purpose is to generate desired footstep positions and orientation and also the desired DCM trajectory. The second layer employs \emph{simplified robot models} to track the desired DCM, CoM and ZMP trajectories. Finally, the third control layer is given by the \emph{whole-body QP} block. It has the main purpose of ensuring the tracking of the desired feet positions and orientations and also the CoM trajectories. 
This, differently from the previous control layer, exploits whole-body robot kinematic models and feedback.

\subsection{Trajectory optimization layer}
\label{sub_sec:Trajectory_optimization}
The main purpose of this layer is to evaluate the desired footstep positions and the desired feet and DCM trajectories.  
\par
To plan the desired footstep positions, the humanoid robot is approximated as an unicycle \cite{Dafarra2018}. The feet are represented by the unicycle wheels, and the footsteps can be obtained through sampling of the unicycle trajectories. 
In particular, given the finite set of unicycle positions, we can sample particular feet positions. 
Each position is associated with a time instant $k$. This time instant is considered as the foot impact time $t_{imp}$. Thus, the impact time can be used as a decision variable, which allows us to select the feet position and avoid fast/slow step duration and long/short step length.
Furthermore, the footsteps are planned to avoid excessive rotation between the two feet positions that could be unfeasible
because of joint limits.
Once the footsteps are planned, the desired feet trajectory is obtained by cubic spline interpolation.
\par
The DCM is chosen so as to satisfy the following time evolution:
\begin{equation}
\label{eq:dcm_solution_ios}
\xi = r^{zmp} + e ^{\omega t} (\xi_0 - r^{zmp}),
\end{equation}
where $\xi_0$ is the initial position of the DCM, $r^{zmp}$ is the position of the ZMP (placed on the center of the stance foot) and $t$ has to belong to the step domain $t \in [0, \; t^{step}_i]$ where $t^{step}_i$ is the duration of the $i$-th step.
Assuming that the final position of the DCM coincides with the ZMP at last step (i.e. $\xi_{N-1}^{eos} = r_{N}^{zmp}$), \eqref{eq:dcm_solution_ios} can be used to find the desired DCM position at the end of each step \cite{Englsberger2014}
\begin{equation}
\label{eq:dcm_eos_evaluation}
\begin{cases}
\xi_{N-1}^{eos} = r_{N}^{zmp} \\
\xi_{i-1}^{eos} = \xi_{i}^{ios} = r_{i}^{zmp} + e^{-\omega t^{step}_i} (\xi_i^{eos} - r_i^{zmp}),
\end{cases}
\end{equation}
where $\xi_{i}^{ios}$ and $\xi_{i}^{eos}$ are respectively the desired DCM initial and final positions for the $i$-th step.
\par
By substitution of \eqref{eq:dcm_eos_evaluation} into \eqref{eq:dcm_solution_ios}, one obtains the reference DCM trajectory:
\begin{equation}
\label{eq:dcm_solution_eos}
\xi_i(t) = r^{zmp}_i + e ^{\omega (t - t^{step}_i)} (\xi^{eos}_i - r^{zmp}_i).
\end{equation}
The DCM velocity can be easily evaluated via differentiation of \eqref{eq:dcm_solution_eos}:
$\dot{\xi}_i(t) =\omega e ^{\omega (t - t^{step}_i)} (\xi^{eos}_i - r^{zmp}_i)$.
In light of the above, the DCM trajectory along the walking pattern can be computed recursively.
The presented planning method is very powerful and it allows us to generate the desired DCM trajectory in real-time. Nevertheless, it has the main limitation of taking into account single support phases only. Indeed, by considering instantaneous transitions between two consecutive single support phases, the ZMP reference is discontinuous.
This leads to the discontinuity of the external forces and also of the desired joint torques.
The development of a DCM trajectory generator that handles non-instantaneous transitions between two single support phases becomes pivotal~\cite{Englsberger2014}.
\par
Since the ZMP is related to the DCM position and velocity, namely $r^{zmp} = \xi - \dot{\xi}/\omega$, a DCM reference trajectory with continuous derivative ($C^1$ continuity) guarantees a continuous ZMP trajectory.
This can be easily ensured using a third order polynomial to smooth the edges of the DCM reference trajectory evaluated using the exponential technique \cite{Englsberger2014}:
\[
\xi^{DS} = a_3 t^3 + a_2 t^2 + a_1 t + a_0,
\]
where the parameters $a_i$ for $i=0:3$ have to be chosen in order to satisfy the velocity and position boundary conditions, namely:
\[
\begin{cases}
\xi^{DS_i}_i = r^{zmp}_{i-1} + e ^{\omega (t^{DS_i}_{i-1} - t^{step}_{i-1})} (\xi^{eos}_{i-1} - r^{zmp}_{i-1}) \\
\dot{\xi}^{DS_i}_i = \omega e ^{\omega (t^{DS_i}_{i-1} - t^{step}_{i-1})} (\xi^{eos}_{i-1} - r^{zmp}_{i-1}) \\
\xi^{DS_e}_i = r^{zmp}_i + e ^{\omega (t^{DS_e}_i - t^{step}_i)} (\xi^{eos}_i - r^{zmp}_i) \\
\dot{\xi}^{DS_e}_i = \omega  e ^{\omega (t^{DS_e}_i - t^{step}_i)} (\xi^{eos}_i - r^{zmp}_i).
\end{cases}
\]
Here $\xi^{DS_i}$ and $\dot{\xi}^{DS_i}$ are the desired DCM position and velocity at the beginning of the double support phase,  $\xi^{DS_e}$ and $\dot{\xi}^{DS_e}$ are the desired DCM position and velocity at the end of the double support phase, $t^{DS_i}$ and
$t^{DS_e}$ are the initial and final instant of the double support phase.
\par
\begin{figure}[tpb]
\centering
\includegraphics[angle =-90, scale=0.5]{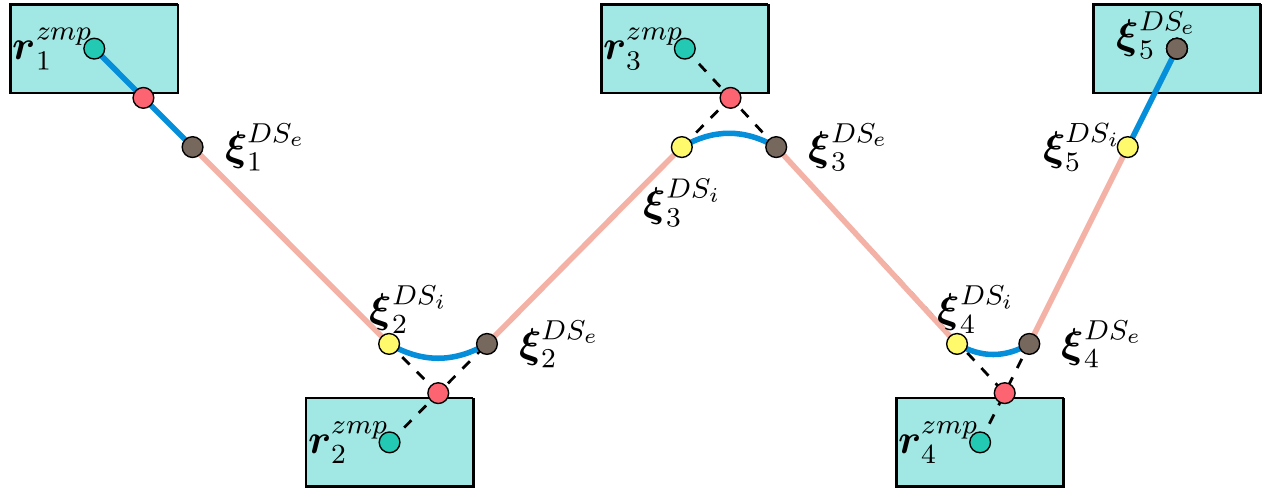}
 \caption{Planning of 2D-DCM on a flat terrain. The single support trajectories are represented by orange segments. The double support trajectories are represented by light blue curves.} \label{fig:dcm_ds}
    \vskip-0.5cm
\end{figure}

In Fig. \ref{fig:dcm_ds} the whole DCM trajectory is shown. During the single support phase (orange segment) the exponential interpolation technique \eqref{eq:dcm_solution_eos} is used; while during the double support phase (light blue curves) the trajectory is obtained using the polynomial technique described above.

\subsection{Simplified model control layer}
As discussed in Sec. \ref{sub_sec:simplified_models}, the simplified model~\eqref{eq:simplified-model} shows that the CoM asymptotically converges to a constant DCM, while the DCM has an unstable first order dynamics.
Then, we present in this section control laws that aim at stabilizing the DCM dynamics only assuming the
$r^{zmp}$ as control input.

This stabilization problem has been tackled by designing and comparing \emph{instantaneous} and \emph{ predictive} (MPC) control laws.
Differently from the MPC, the instantaneous control law does not solve any optimization
problem and it uses only
the current desired and actual position of the DCM to evaluate its output.
\par

\subsubsection{DCM instantaneous control}
\label{instantaneous-control}

Differently from \cite{Englsberger2015}, we propose an instantaneous control law that does not perform a cancellation of the unstable system dynamics, which, in practice, is never perfect and may lead to system instabilities. More precisely, we choose
\begin{equation}
\label{eq:reactive_dcm}
\begin{split}
r^{zmp}_{ref} {=} \xi_{ref} {-} \frac{\dot{\xi}_{ref}}{\omega} {+} K^{\xi}_{p} (\xi {-} \xi_{ref}) 
+ K^{\xi}_{i} \int{(\xi {-} \xi_{ref}) \diff t} ,
\end{split}
\end{equation}
where  $K^{\xi}_{p}> I_2$ and $K^{\xi}_{i} > 0_2$
\par
Applying the control input \eqref{eq:reactive_dcm} to  system \eqref{eq:dcm_dynamics} leads to the following closed loop dynamics:
\begin{equation}
\label{eq:reactive_cl}
\begin{split}
(\dot{\xi} {-} \dot{\xi}_{ref}) {=} \omega(I_2 {-} K^{\xi}_{p})  (\xi {-} \xi_{ref})
{-}\omega K^{\xi}_{i}  \int{(\xi {-} \xi_{ref}) \diff t}.
\end{split}
\end{equation}
It is easy to show the DCM error and its integral converge asymptotically to zero.
The above law~\eqref{eq:reactive_dcm} is very simple to implement and guarantees the tracking of the
desired DCM. 
The main limitation is that may not ensure the feasibility of the gait since the position of the ZMP may exit the support polygon.
Furthermore, the above instantaneous control law does not take into account the future planned desired DCM trajectory.

\subsubsection{DCM predictive control}
\label{predictive-control}
In order to overcome these issues,  a model predictive controller can be designed~\cite{Krause2012}.
\par
In the MPC framework, the DCM dynamics \eqref{eq:dcm_dynamics} is used as a prediction model and it
is discretized supposing piecewise constant ZMP trajectories:
\begin{equation}
\label{eq:dcm_discrete}
\begin{split}
\xi_{k+1} &{=} F \xi_k + G r_k^{zmp} \\
&{=}
\begin{bmatrix}
e^{\omega T} & 0 \\
0 & e^{\omega T}
\end{bmatrix} \xi_k {+}
\begin{bmatrix}
1 - e^{\omega T} & 0\\
0 & 1 - e^{\omega T}
\end{bmatrix}
r_k^{zmp}.
\end{split}
\end{equation}
In order to ensure that the stance foot does not rotate around one of its edges, the desired ZMP must not exit the support polygon \cite{Vukobratov2004}. This is verified by means of a set of linear inequality constraints:
\begin{equation}
\label{eq:inequality_constraints}
A_{c_k} r^{zmp}_k \le b_{c_k},
\end{equation}
where $A_{c_k}$ and $b_{c_k}$ are time variant and their dimension depends on the type of support.
\par
The cost function shall then ensure the tracking of the desired trajectory.
In  particular, we choose the following cost:
\begin{IEEEeqnarray}{LL}
\IEEEyesnumber \phantomsection
\label{eq:mpc_const_function}
J_k = & \sum\limits_{j=k}^{N + k-1} ({\xi}_j-{\xi}_j^{ref})^\top Q ({\xi}_j-{\xi}_j^{ref}) + \IEEEyessubnumber \label{eq:mpc_cost_state}\\
&  ({r}^{zmp}_j - {r}^{zmp}_{j-1})^\top R ({r}^{zmp}_j - {r}^{zmp}_{j-1}) + \IEEEyessubnumber \label{eq:mpc_cost_input}\\
&({\xi}_{k+N}-{\xi}_{k+N}^{ref})^\top Q_N ({\xi}_{k+N}-{\xi}_{k+N}^{ref})  \IEEEyessubnumber \label{eq:mpc_cost_FinalTerm}
\end{IEEEeqnarray}
where $Q$, $Q_N$ and $R$ are positive $2 \times 2$ symmetric matrices and $N$ is length of the preview
window.
\par
The terms \eqref{eq:mpc_cost_FinalTerm} \eqref{eq:mpc_cost_state} guarantee the tracking of the
given reference trajectory, while \eqref{eq:mpc_cost_input} is added for obtaining a
smoother ZMP trajectory.
Even if the cost function $J_k$ is time-dependent, it is always positive and convex.
\par
The MPC problem can be summarized as follows:
\begin{IEEEeqnarray}{LCL}
\label{MPC_solution}
r_k^{{zmp}^{*}} =& \argmin\limits_{\substack{\xi_k, \dots, \xi_{k + N}\\  r^{zmp}_k, \dots, r^{zmp}_{k + N-1}}} & J_k  \\ \nonumber \\
&\text{s.t.}   &{\xi}_{i+1} = F {\xi}_i + G {r}^{zmp}_i \nonumber \\
& &  A_{c_k} {r}^{zmp}_{k} \leq {b}_{c_k} \nonumber \\
& & {\xi}_k = \bar{{\xi}} \nonumber \\
& & {r}^{zmp}_{k-1} = \bar{{r}}^{zmp}, \nonumber
\end{IEEEeqnarray}
where $i$ satisfies $k\le i \le N + k - 1$, $\bar{{r}}^{zmp}$ is the desired ZMP
computed at the previous control iteration and $\bar{{\xi}}$ is the
estimated DCM position.
\par
Since the cost function is a quadratic positive function and the constraints are linear, the optimal control problem is quadratic and it can be converted into a strictly convex quadratic programming
problem (QP) of the form:
\begin{IEEEeqnarray*}{CLLL}
\min\limits_{w_k} & \IEEEeqnarraymulticol{3}{C}{\frac{1}{2} w_k^\top H_k w_k + g_k^\top w_k}\\
\text{s.t.} & A_{c_k}^i w_k &\le &b_{c_k}^i\\
& A_{c_k}^e w_k &= &b_{c_k}^e.
\end{IEEEeqnarray*}
Here the optimization variables are stacked inside the vector $w_k = \begin{bmatrix}
\xi_k ^\top & \hdots & \xi_{k + N} ^\top &
r_{k}^{zmp ^\top} & \hdots & r_{k + N - 1} ^{zmp ^\top}
\end{bmatrix}^\top$.
The Hessian matrix $H_k$ and the gradient vector $g_k$ can be obtained from \eqref{eq:mpc_const_function}.
The inequality constraint matrix $A_{c_k}^i$ and vector $b_{c_k}^i$ embed the ZMP constraint
\eqref{eq:inequality_constraints}.
While the equality constraints, $A_{c_k}^e$ and $b_{c_k}^e$ are determined using the prediction
model \eqref{eq:dcm_discrete}.

\subsubsection{ZMP-CoM Controller}
\label{ZMP-CoM-Controller}
Independently from the chosen DCM controller, namely either the controller in Sec.~\ref{instantaneous-control} or in~\ref{predictive-control}, one obtains a desired ZMP and CoM position and velocity to be stabilised.  As consequence, another control loop is needed after the DCM controller. In this paper, we choose the proposed control law \cite{Choi2007}, i.e.:
\begin{equation}
\label{eq:zmp_controller}
\dot{x}^* = \dot{x}_{ref} - K_{zmp}(r^{zmp}_{ref} - r^{zmp}) + K_{com} (x_{ref} - x),
\end{equation}
where $K_{com} > \omega I_2$  and $0_2 < K_{zmp} < \omega I_2.$

\subsection{Whole-body QP control layer}

The main control objective for the whole-body QP control layer is to stabilise some robot kinematic quantities by using the entire robot body. We here choose to track the position
of the CoM, the torso orientation, and the left and right feet position and orientation.
To do so, we use a stack of tasks formulation. The tracking of the feet poses and of the CoM position is considered as high priority tasks (hard constraint), while the torso orientation is considered as a low priority task (soft constraint). Furthermore, a postural condition is also added as a low-priority task.
Then, the following cost function is defined:
\begin{IEEEeqnarray}{LL}
\IEEEyesnumber \phantomsection
\label{eq:velocity_control_cost}
f(\nu) {=} \frac{1}{2} [ & (v ^ * _ {\mathcal{T}}{-}J_{\mathcal{T}} \nu)^\top
K_{\mathcal{T}} (v ^ * _ {\mathcal{T}}{-}J_{\mathcal{T}} \nu) {+} (\dot{s} {-} \dot{s}^ * ) ^\top
\Lambda (\dot{s} {-} \dot{s}^*)].   \label{eq:velocity_control_regularization_cost}  \IEEEeqnarraynumspace
\end{IEEEeqnarray}
with $K_{\mathcal{T}} > 0$  and $
v^*_{\mathcal{T}} = -K _{\omega  _{\mathcal{T}}} \text{sk}(\prescript{\mathcal{I}}{}{R} _{\mathcal{T}} {\prescript{\mathcal{I}}{}{R}_{\mathcal{T}} ^{*}}^{\top})^{\vee},
$
where $K _{\omega  _{\mathcal{T}}} > 0$. This latter control law guarantees almost-global stability and convergence of ${}^\mathcal{I}R _{\mathcal{T}}$ to ${}^\mathcal{I}R _{\mathcal{T}}^*$  \cite{Olfati-Saber:2001:NCU:935467}.
\par
The postural task \eqref{eq:velocity_control_regularization_cost}, with $\Lambda> 0$, is achieved by
asking for a desired joints velocity that depends on the error between the desired and measured
joints position 
\begin{equation}
\label{eq:regularization_term}
\dot{s}^ * = -K_{s} (s - s^d),    
\end{equation}
where $K_{s}$
is a positive definite matrix.
\par
Concerning the hard constraints, we have:
\begin{equation}
\label{eq:velocity_control_jacobians}
J_{\mathcal{C}}(\nu) \nu = v^ * _ {\mathcal{C}}, \quad
J_{\mathcal{LF}}(\nu) \nu = v^ * _ {\mathcal{LF}}, \quad
J_{\mathcal{RF}}(\nu) \nu = v^ * _ {\mathcal{RF}},
\end{equation}
where $v^*_{\mathcal{C}}$ is the linear velocity of the CoM, $v^*_{\mathcal{LF}}$ and $ v^*_{\mathcal{RF}}$ are respectively the
desired left foot and right foot velocities. More specifically $v^*_{\mathcal{\#F}}$, where
$\#=[\mathcal{R}, \mathcal{L}]$, is chosen as:
\begin{equation}
\label{feetVelocitiesStar}
v^*_ {\# \mathcal{F}} = \prescript{\mathcal{I}}{}{\dot{p}}^*_{\# \mathcal{F}} -
\begin{bmatrix}
K^p _{x _{\# f}} e^p_{\#f}
+ K^i _{x _{\# f}}\int{ e^p_{\#f} \diff t}\\
K _{\omega  _{\# f}} \text{sk}(\prescript{\mathcal{I}}{}{R}_{\#\mathcal{F}} \prescript{\mathcal{I}}{}{R} _{\# \mathcal{F}} ^{*^\top} )^{\vee}
\end{bmatrix}.
\end{equation}
Here $e^p_{\#f} = \prescript{\mathcal{I}}{}{p}_{\# \mathcal{F}}
- \prescript{\mathcal{I}}{}{p}^*_{\# \mathcal{F}}$, while the gains $K^p _{x _{\# f}}$, $K^i _{x _{\# f}}$, $K _{\omega _{\# f}}$ are positive definite matrices. 
\par
Finally, the desired velocity of the CoM $v^ * _{\mathcal{C}}$ is chosen as:
\[
v^*_{\mathcal{C}} = \dot{x}^* - K^p _{\mathcal{C}}(x - x^*) - K^i _{\mathcal{C}}\int{ x - x^*  \diff t},
\]
where the gain matrices are positive definite positive $\dot{x}^*$ is the output of the ZMP-CoM \eqref{eq:zmp_controller} controller and $x^*$ is the integrated signal. Finally, we add constraints
on the maximum  velocity 
\begin{equation}
\label{eq:ik_jacobian_limits}
\dot{s}^- \le \dot{s} \le \dot{s}^+.
\end{equation}
\par
The above hierarchical control objectives can be cast into a whole-body optimization problem:
\begin{IEEEeqnarray}{LCL}
\nu ^* =& \argmin\limits_{\nu}  \frac{1}{2} [&(v^* _ {\mathcal{T}}-J_{\mathcal{T}} \nu)^\top
K_{\mathcal{T}} (v^*_{\mathcal{T}}-J_{\mathcal{T}} \nu) + \nonumber \\
& & (\dot{s} - \dot{s}^ * )^\top \Lambda (\dot{s} - \dot{s}^*) ] \label{velocityQP}\\
&\text{s.t.} & \dot{s} = S \nu \nonumber \\
& & \dot{s}^ * = -K_{s} (s - s^d) \nonumber \\
& & J_{\mathcal{C}} \nu = v^ * _ {\mathcal{C}} \nonumber \\
& & J_{\mathcal{LF}} \nu = v^ * _ {\mathcal{LF}} \quad J_{\mathcal{RF}} \nu = v^ * _ {\mathcal{RF}} \nonumber \\
& & \dot{s}^- \le \dot{s} \le \dot{s}^+ \nonumber
\end{IEEEeqnarray}
Since the decision variable is the robot velocity $\nu$ and the body velocity depends, through the
Jacobian matrices, linearly on $\nu$ the optimization problem can be converted to the QP problem of
the form:
\begin{IEEEeqnarray*}{CLL}
\min\limits_{\nu} & \IEEEeqnarraymulticol{2}{C}{\frac{1}{2}\nu ^\top H \nu + g^\top \nu} \\
\mbox{s.t.} & A_{c} \nu = b_{c}\\
&\dot{s}^- \le \dot{s} \le \dot{s}^+.
\end{IEEEeqnarray*}
The Hessian matrix $H$ and the gradient vector $g$ are evaluated from \eqref{eq:velocity_control_cost}.
The constraint matrix and vector $A_c$ and  $b_c$ are obtained from \eqref{eq:velocity_control_jacobians}.
Using this formulation the optimization problem can be solved using a standard numerical QP solver.

\subsubsection{Position and velocity controlled robot}
\label{subsubsec-pos-vel-control}
It is important to notice that the outcome of~\eqref{velocityQP} is (also) the robot joint velocity. When a robot velocity controller is available, one can set these joint velocities to the low level robot controller. In this case, the  \emph{*} quantities in~\eqref{velocityQP} can be evaluated by using robot sensor feedback, and the robot is said to be \emph{velocity controlled}. On the other hand, if the robot velocity control is not available, one may integrate the outcome of~\eqref{velocityQP} to obtain desired joint position to be set to a low level robot position controller. In this case, the  \emph{*} quantities in~\eqref{velocityQP} can be evaluated by using the desired integrated quantities instead of sensor feedback, and~\eqref{velocityQP} behaves as an inverse kinematics module, and the robot is said to be \emph{position controlled}.

%% file: tex/results.tex
\section{EXPERIMENTAL RESULTS}
\label{sec:EXPERIMENTAL_RESULTS}
In this section, we present experiments obtained with several implementations of the control architecture shown in Fig.~\ref{fig:controller_architecture}. 
We use the iCub humanoid robot \cite{Metta2010} to carry out the experimental activities. Let us recall that iCub is $\SI{104}{\centi \meter}$ tall, with a foot length and width of  $\SI{19}{\centi \meter}$ and $\SI{9}{\centi \meter}$, respectively.

\begin{table}[b]
    \vskip-0.5cm
    \centering
    \caption{Maximum forward straight walking velocities achieved using different implementations of the control architecture.
    }
    \begin{tabular}{cc|c}
         \begin{tabular}{@{}c@{}}Simplified Model \\ Control\end{tabular} &
         \begin{tabular}{@{}c@{}}Whole-Body \\ QP Control\end{tabular} &
         \begin{tabular}{@{}c@{}}Max Straight \\ Velocity (m/s)\end{tabular}\\
        \hline
        Predictive  & Velocity  &  0.19\\
        Predictive  & Position  & 0.20\\
        Instantaneous  & Velocity  &  0.22\\
        Instantaneous  & Position  & 0.41
    \end{tabular}
    \label{tab:max_velocity}
\end{table}

The control architecture runs on the iCub head's computer, namely a 4-th generation Intel \textsuperscript{\tiny\textregistered} Core i7 @ $\SI{1.7}{\giga \hertz}$. In any of its implementations, the architecture takes (in average) less than $\SI{3}{\milli \second}$ for evaluating its outputs. OSQP \cite{osqp} library was used to solve the optimization problems.


Tab.~\ref{tab:max_velocity} summarizes the maximum velocities achieved using the different implementations of the control architecture. In particular, the labels \emph{instantaneous} and \emph{predictive} mean that the associated layer generates its outputs considering inputs and references either at the single time $t$ or for a time window, respectively. The labels, \emph{velocity} and \emph{position} control, instead, mean that the layer outputs are either desired joint velocities or position, respectively -- see Sec.~\ref{subsubsec-pos-vel-control}. 

Let us remark that all the implemented control architectures exploit the controller presented in section~\ref{ZMP-CoM-Controller} to attempt the stabilization of the desired center-of-pressure and desired center-of-mass position and velocity. The performances of this controller much depend on the gains $K_{zmp}$ and $K_{com}$. In particular, we observed that the gains achieving good tracking during  standing and walking were not the same. For this reason,  we implemented a  gain-scheduling technique depending on the robot is walking or standing. The transition between the two sets of gains is smoothed with a minimum jerk trajectory \cite{Pattacini2010a}.


Comparing different control architectures, however, is a far cry from being an easy task. To do so, we decided to perform two main experiments only, which are used as benchmarks for all control architecture implementations. These two experiments represent the maximum robot velocity that has been achieved with all architectures, and the maximum velocity achieved with a specific architecture only -- see Tab.~\ref{tab:max_velocity}. Namely, 
\begin{itemize}
    \item[\textbf{- Experiment 1}] a forward robot speed of $\SI{0.19}{\meter \per \second}$;
    \item[\textbf{- Experiment 2}] a forward robot speed of $\SI{0.41}{\meter \per \second}$.
\end{itemize}

\begin{figure*}[t]
    \centering
    \begin{myframe}{Instantaneous + Position Control}
    \begin{subfigure}[b]{0.329\textwidth}
        \centering
        \includegraphics[width=\textwidth]{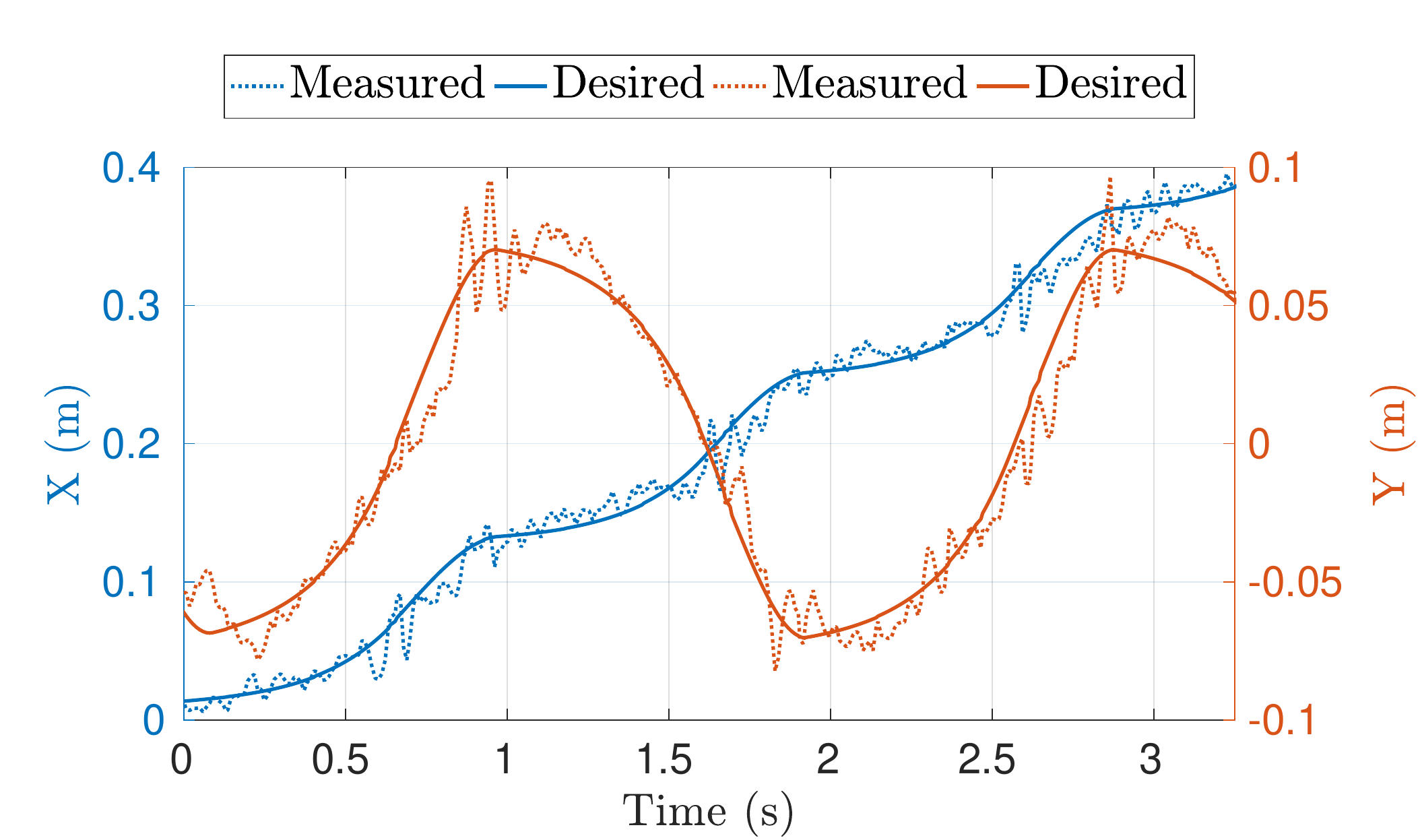}
        \caption{DCM}
        \label{fig:inst_pos-min_vel-dcm}
    \end{subfigure}
    \hfill
    \begin{subfigure}[b]{0.329\textwidth}
        \centering
        \includegraphics[width=\textwidth]{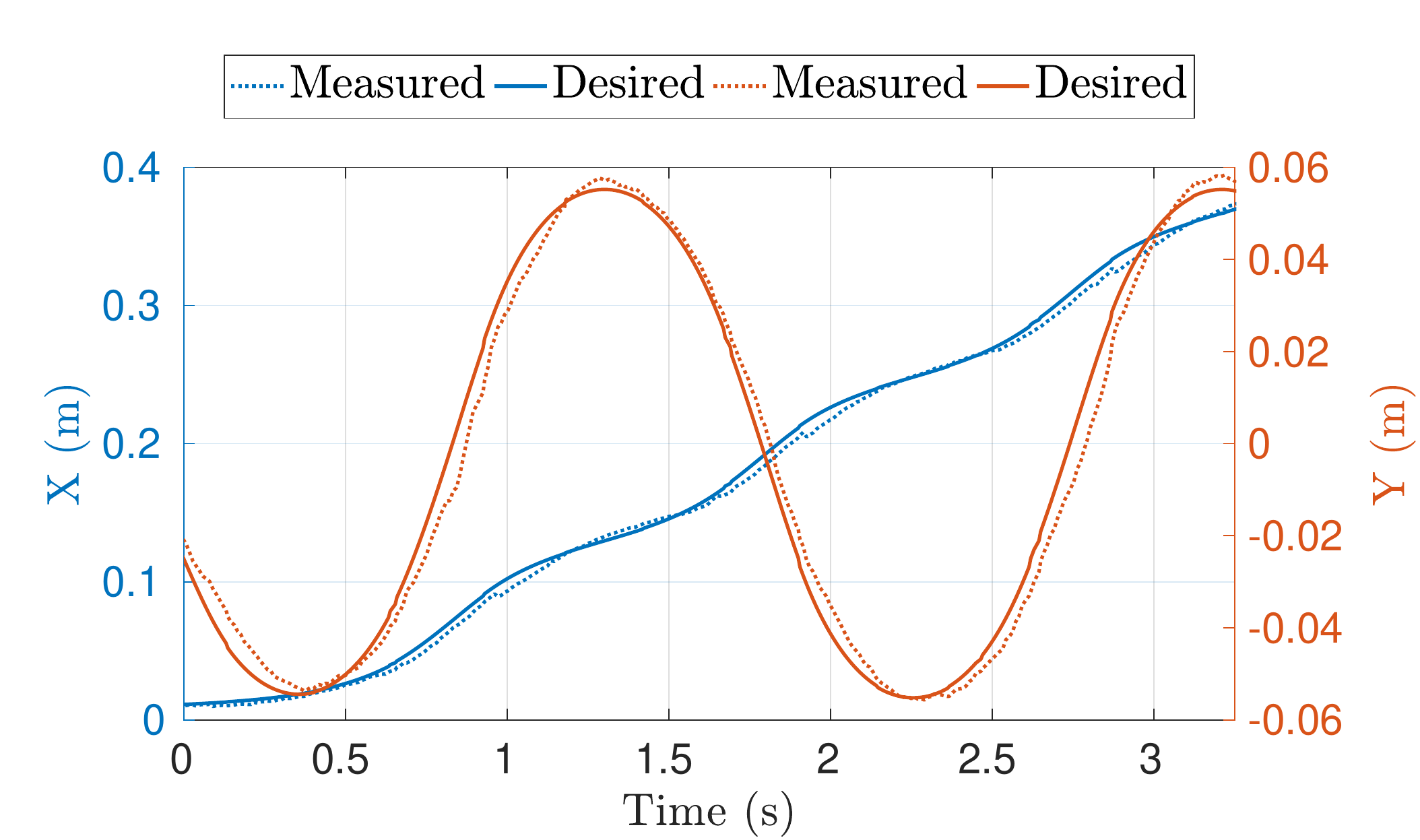}
        \caption{CoM}
        \label{fig:inst_pos-min_vel-com}
    \end{subfigure}
    \hfill
    \begin{subfigure}[b]{0.329\textwidth}
        \centering
        \includegraphics[width=\textwidth]{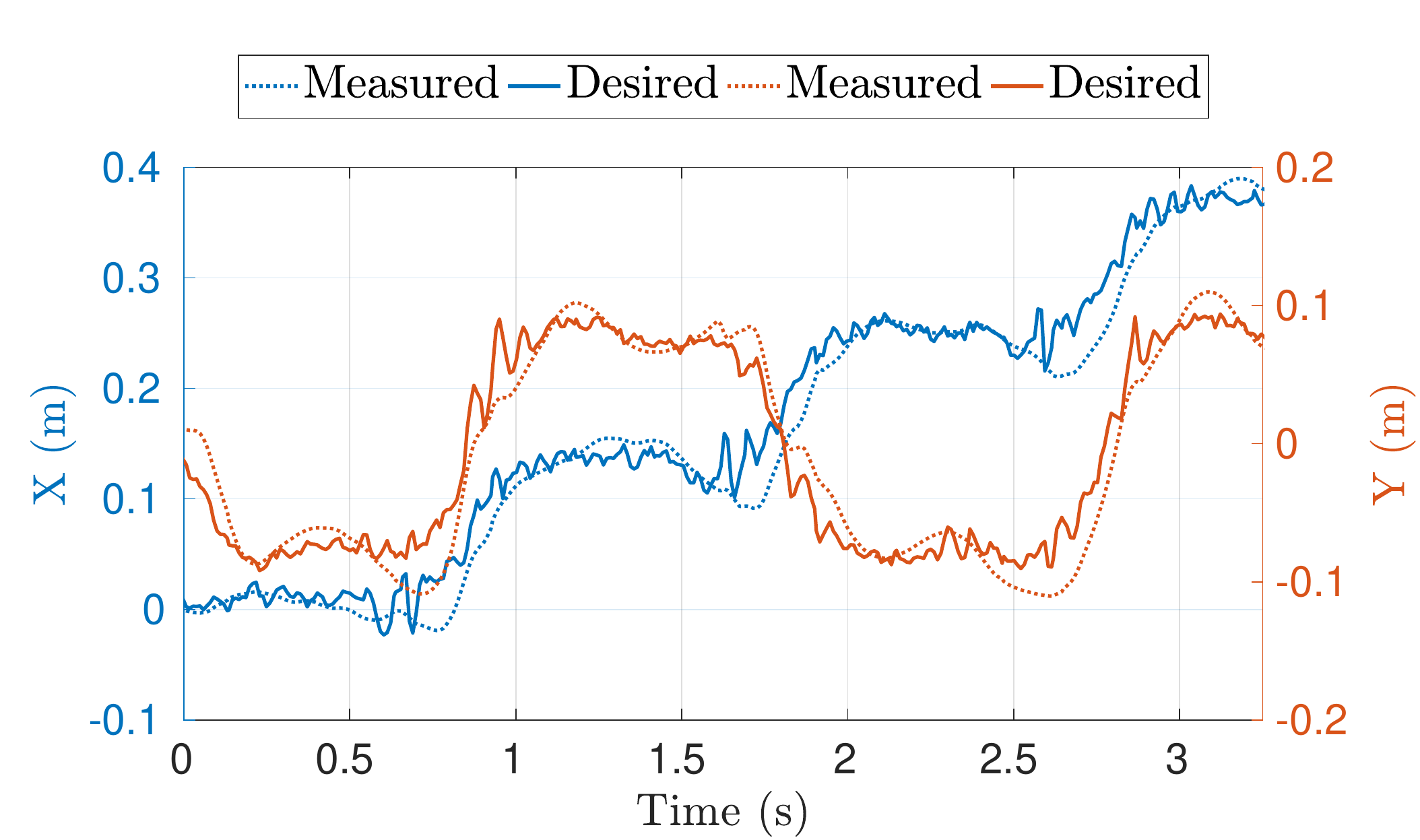}
        \caption{ZMP}
        \label{fig:inst_pos-min_vel-zmp}
    \end{subfigure}
    \end{myframe}
  \vskip-0.5cm
  \hfill
    \begin{myframe}{Predictive + Position Control}
    \begin{subfigure}[b]{0.329\textwidth}
        \centering
        \includegraphics[width=\textwidth]{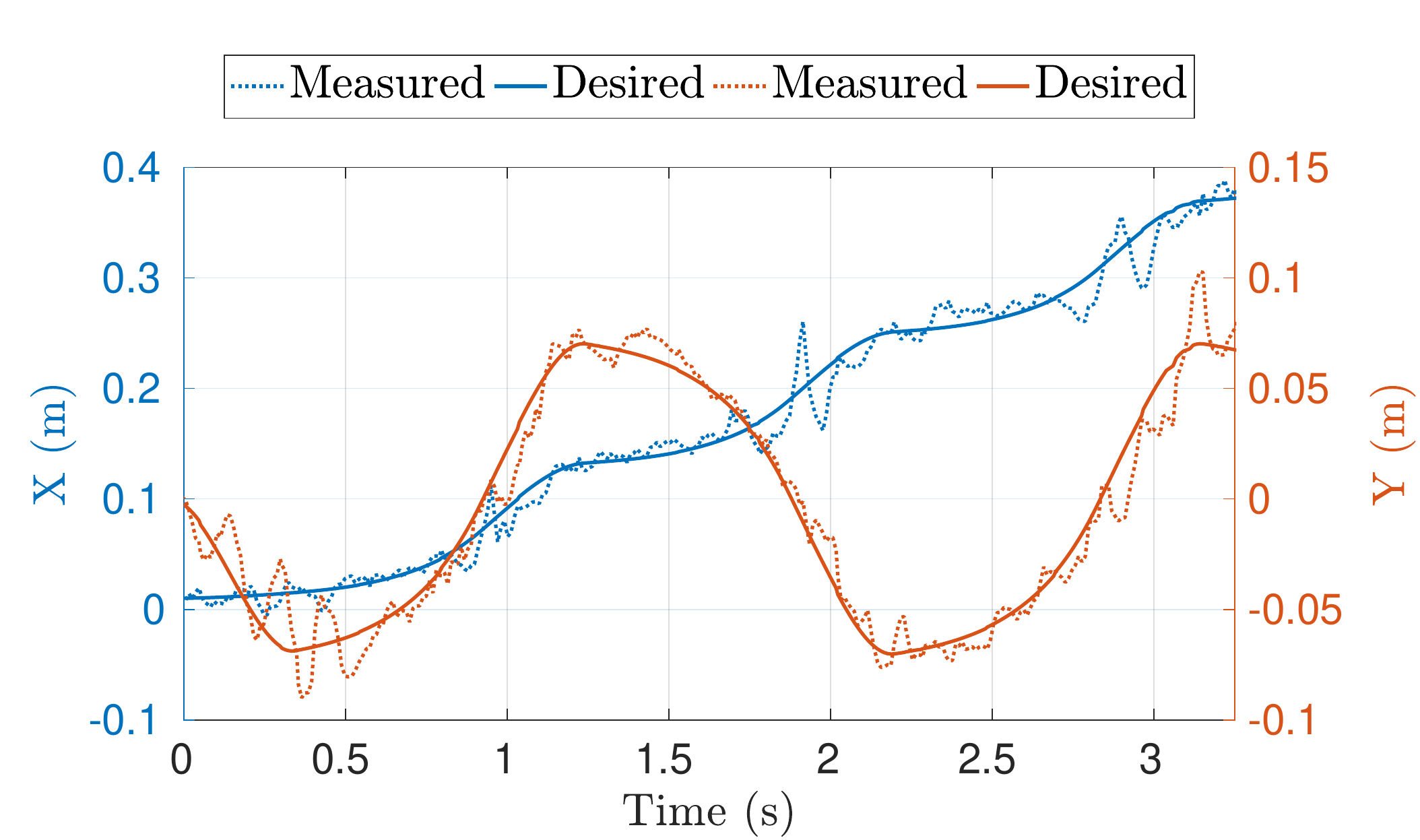}
        \caption{DCM}
        \label{fig:mpc_pos-min_vel-dcm}
    \end{subfigure}
    \hfill
    \begin{subfigure}[b]{0.329\textwidth}
        \centering
        \includegraphics[width=\textwidth]{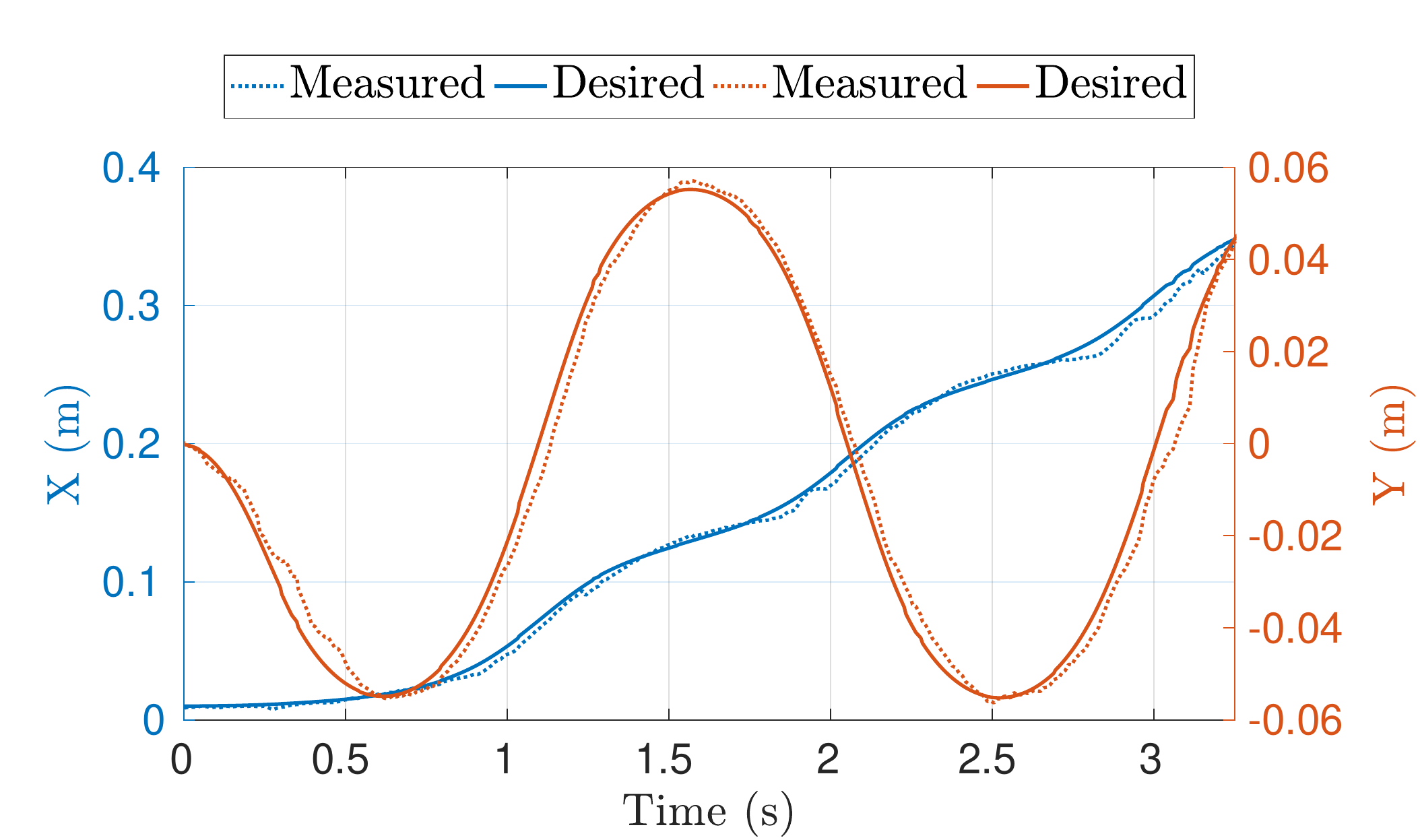}
        \caption{CoM}
        \label{fig:mpc_pos-min_vel-com}
    \end{subfigure}
         \begin{subfigure}[b]{0.329\textwidth}
        \centering
        \includegraphics[width=\textwidth]{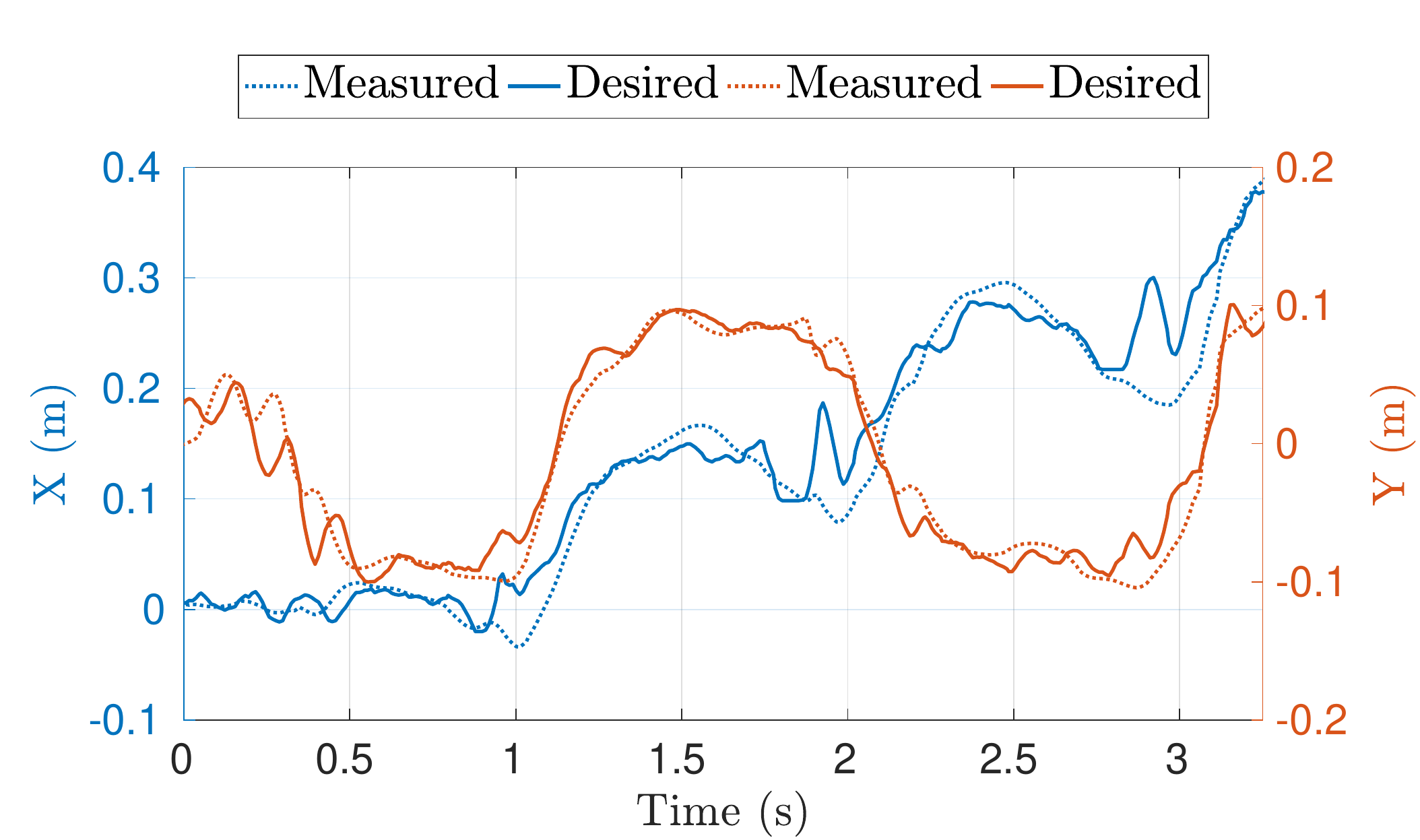}
        \caption{ZMP}
        \label{fig:mpc_pos-min_vel-zmp}
    \end{subfigure}
    \end{myframe}
    \caption{ Tracking of the DCM (a), CoM (b) and ZMP (c) using the instantaneous controller with the whole-body controller as position control. Tracking of the  DCM (d), CoM (e) and ZMP (f) using the MPC and the whole-body controller as position control. Walking velocity:  $\SI{0.19}{\meter \per \second}$.}
\end{figure*}
\begin{figure*}[t]
     \vspace*{-0.1cm}
    \begin{myframe}{Instantaneous + Position Control}
        \begin{subfigure}[b]{0.329\textwidth}
        \centering
        \includegraphics[width=\textwidth]{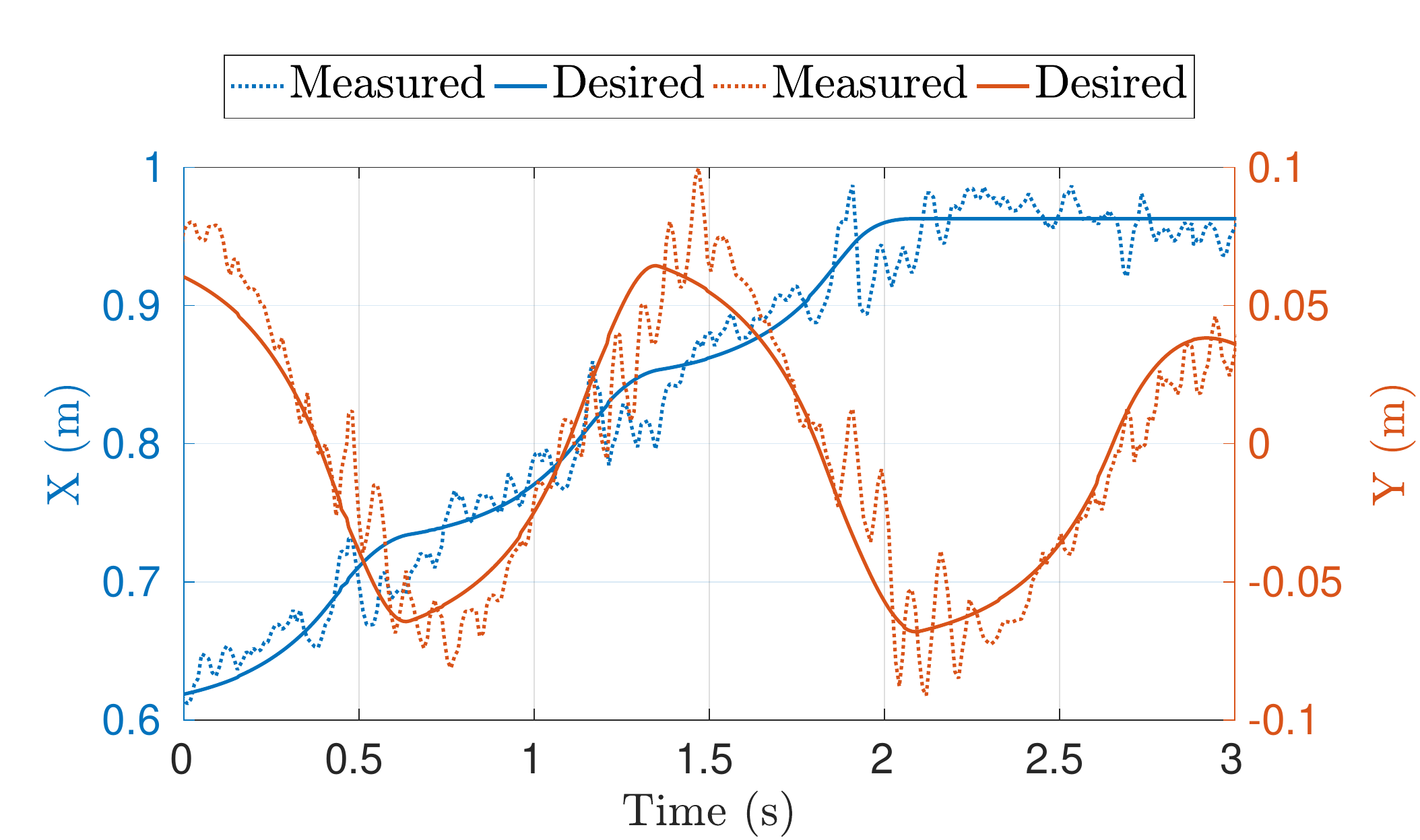}
        \caption{DCM}
        \label{fig:inst_pos-max_vel-dcm}
    \end{subfigure}
    \hfill
     \begin{subfigure}[b]{0.329\textwidth}
        \centering
        \includegraphics[width=\textwidth]{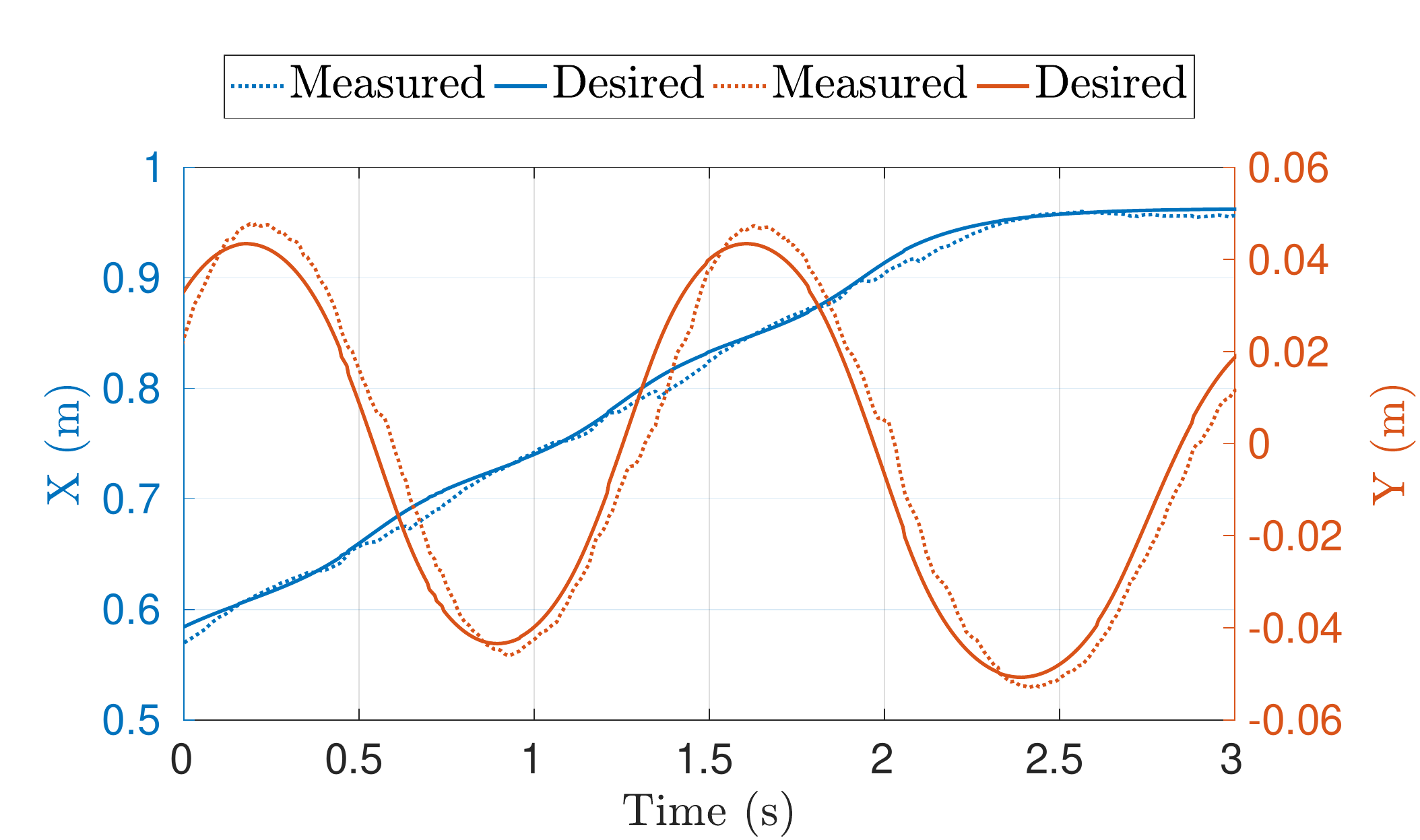}
        \caption{CoM}
        \label{fig:inst_pos-max_vel-com}
    \end{subfigure}
    \hfill
    \begin{subfigure}[b]{0.329\textwidth}
        \centering
        \includegraphics[width=\textwidth]{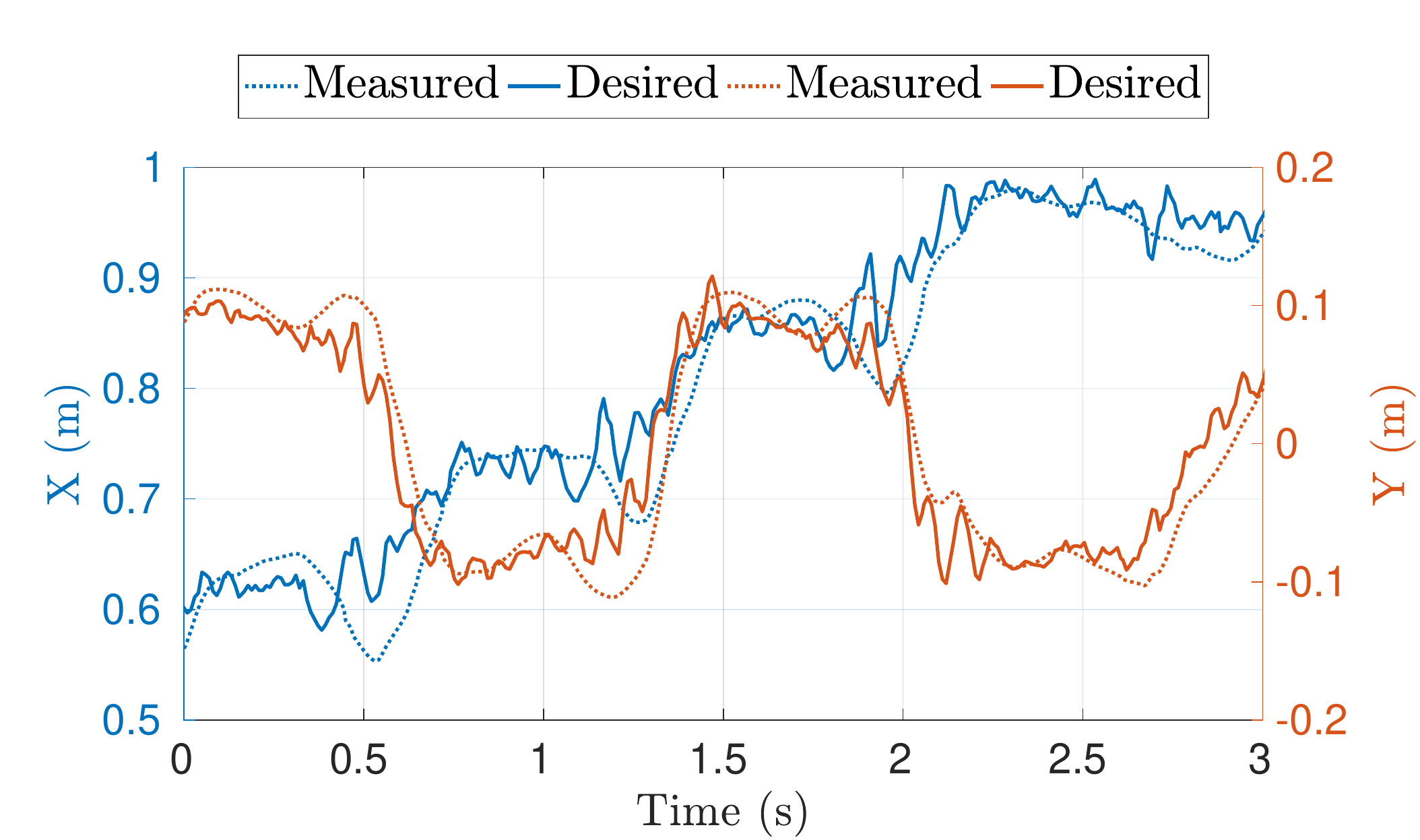}
        \caption{ZMP}
        \label{fig:inst_pos-max_vel-zmp}
    \end{subfigure}
    \end{myframe}
    \vskip-0.5cm
    \hfill
    \begin{myframe}{Predictive + Position Control}
    \begin{subfigure}[b]{0.329\textwidth}
        \centering
        \includegraphics[width=\textwidth]{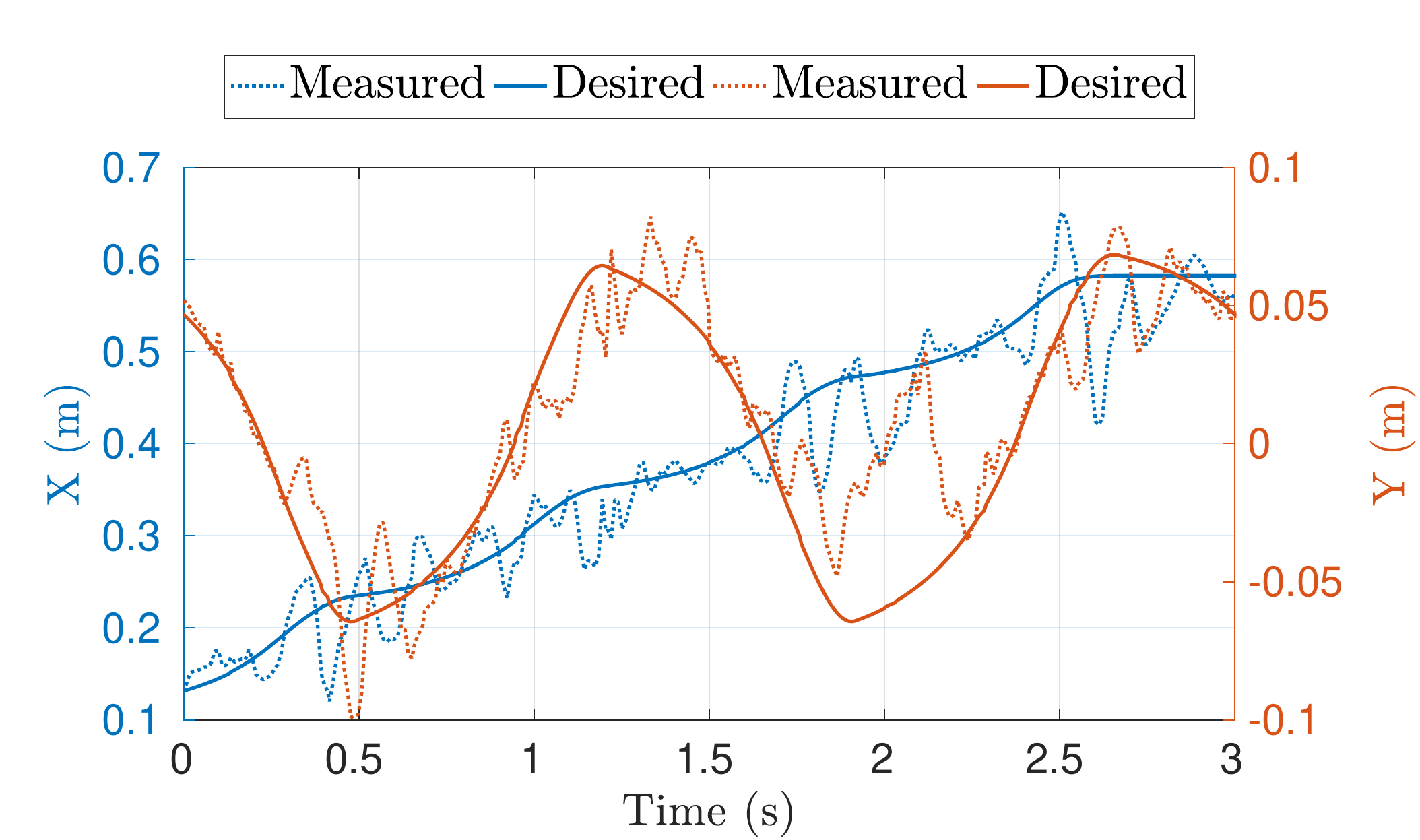}
        \caption{DCM}
        \label{fig:mpc_pos-max_vel-dcm}
    \end{subfigure}
    \hfill
     \begin{subfigure}[b]{0.329\textwidth}
        \centering
        \includegraphics[width=\textwidth]{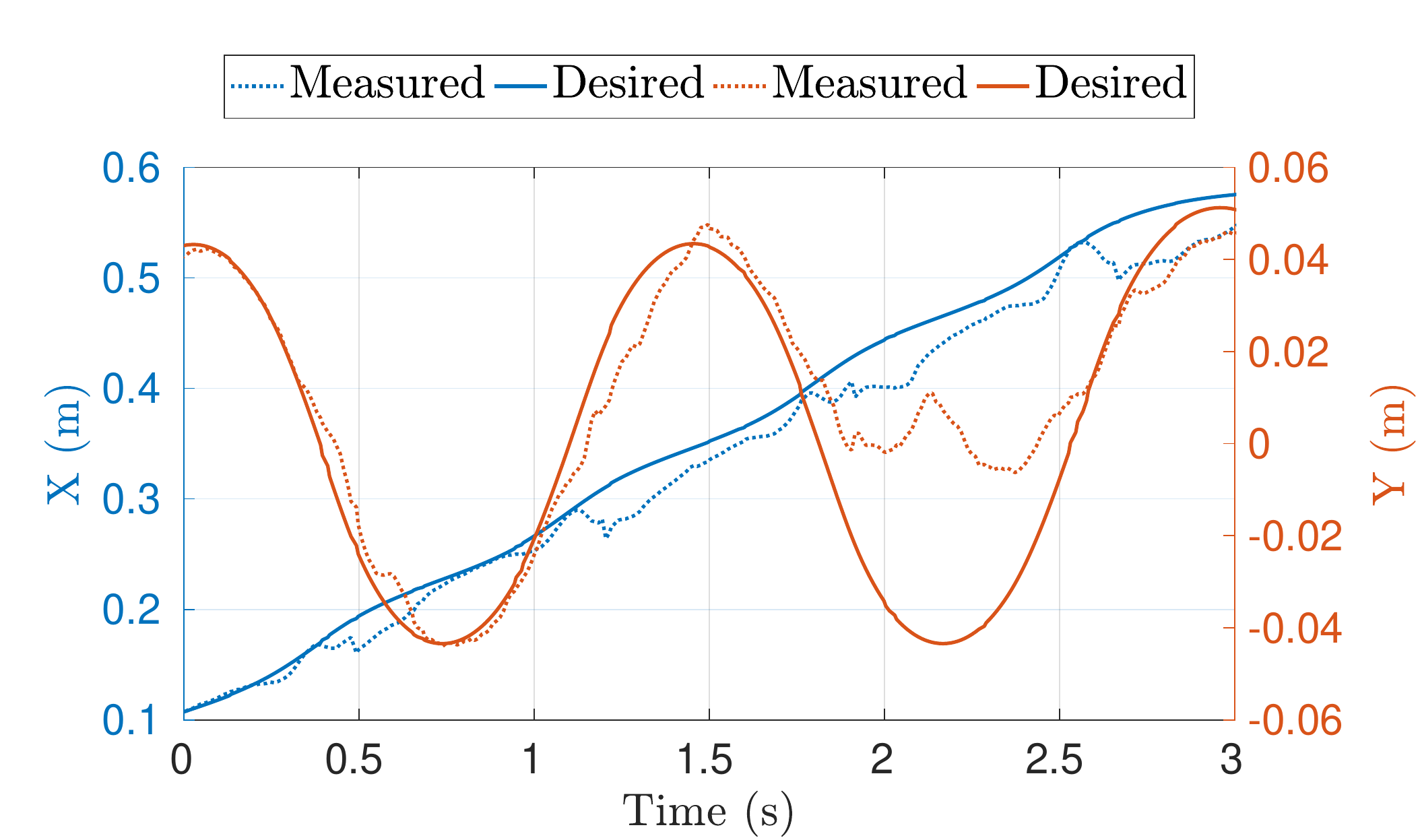}
        \caption{CoM}
        \label{fig:mpc_pos-max_vel-com}
    \end{subfigure}
    \hfill
    \begin{subfigure}[b]{0.329\textwidth}
        \centering
        \includegraphics[width=\textwidth]{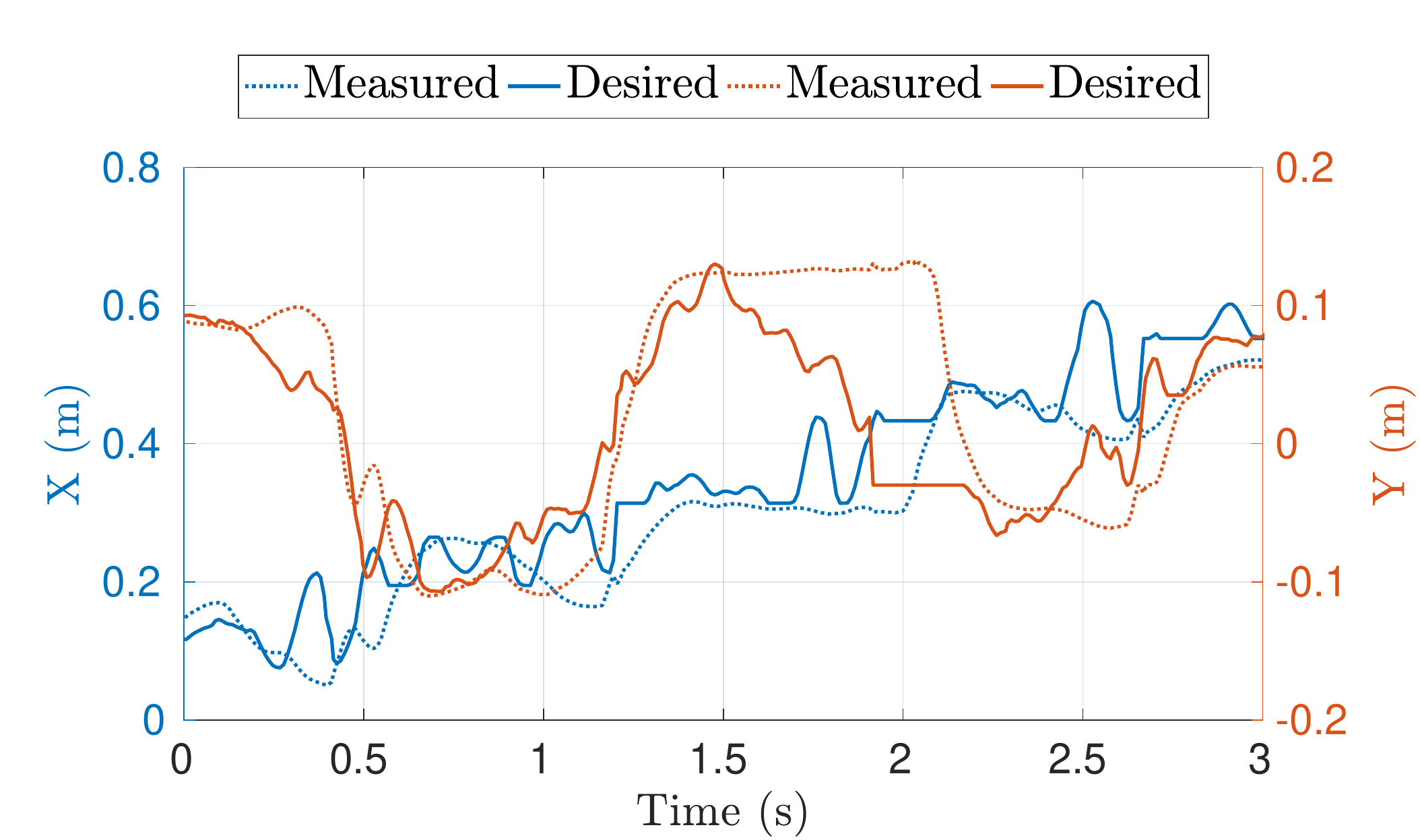}
        \caption{ZMP}
        \label{fig:mpc_pos-max_vel-zmp}
    \end{subfigure}
    \end{myframe}
    \caption{Tracking of the DCM (a), CoM (b) and ZMP (c) with the instantaneous and whole-body QP control as position. Tracking of the  DCM (d), CoM (e) and ZMP (f) with the predictive and whole-body QP control as position control. At $t\approx \SI{2}{\second}$, the robot falls down.  Walking velocity: $\SI{0.41}{\meter \per \second}$.}
    \vskip-0.5cm
\end{figure*}
\begin{figure*}[t]
    \centering
    \begin{subfigure}[b]{\columnwidth}
        \begin{myframe}{Instantaneous + Position Control}
            \begin{subfigure}[t]{0.49\columnwidth}
            \centering
            \includegraphics[width=\textwidth]{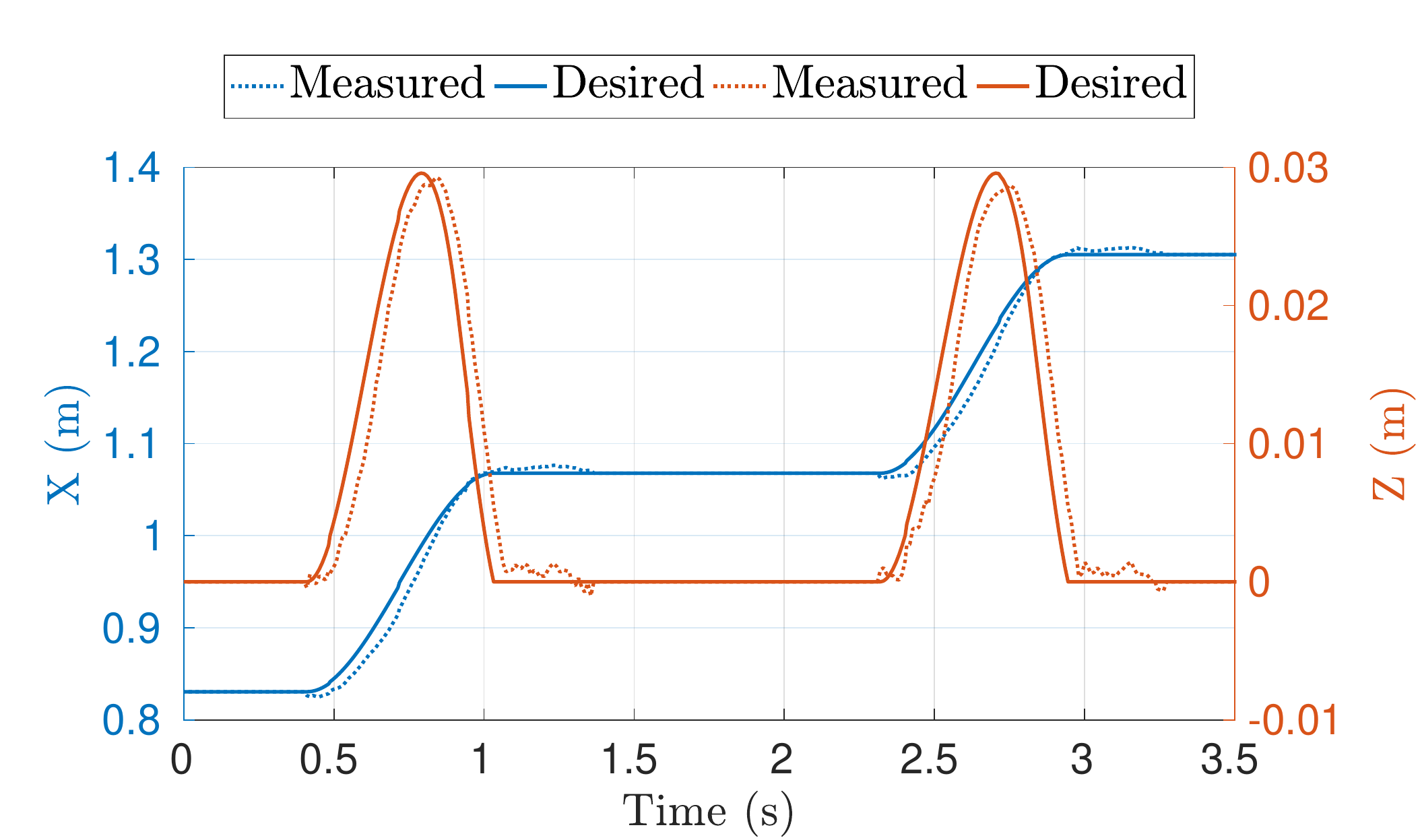}
            \caption{ }
            \label{fig:inst_pos-min_vel-lf}
            \end{subfigure}
            \begin{subfigure}[t]{0.49\columnwidth}
            \centering
            \includegraphics[width=\textwidth]{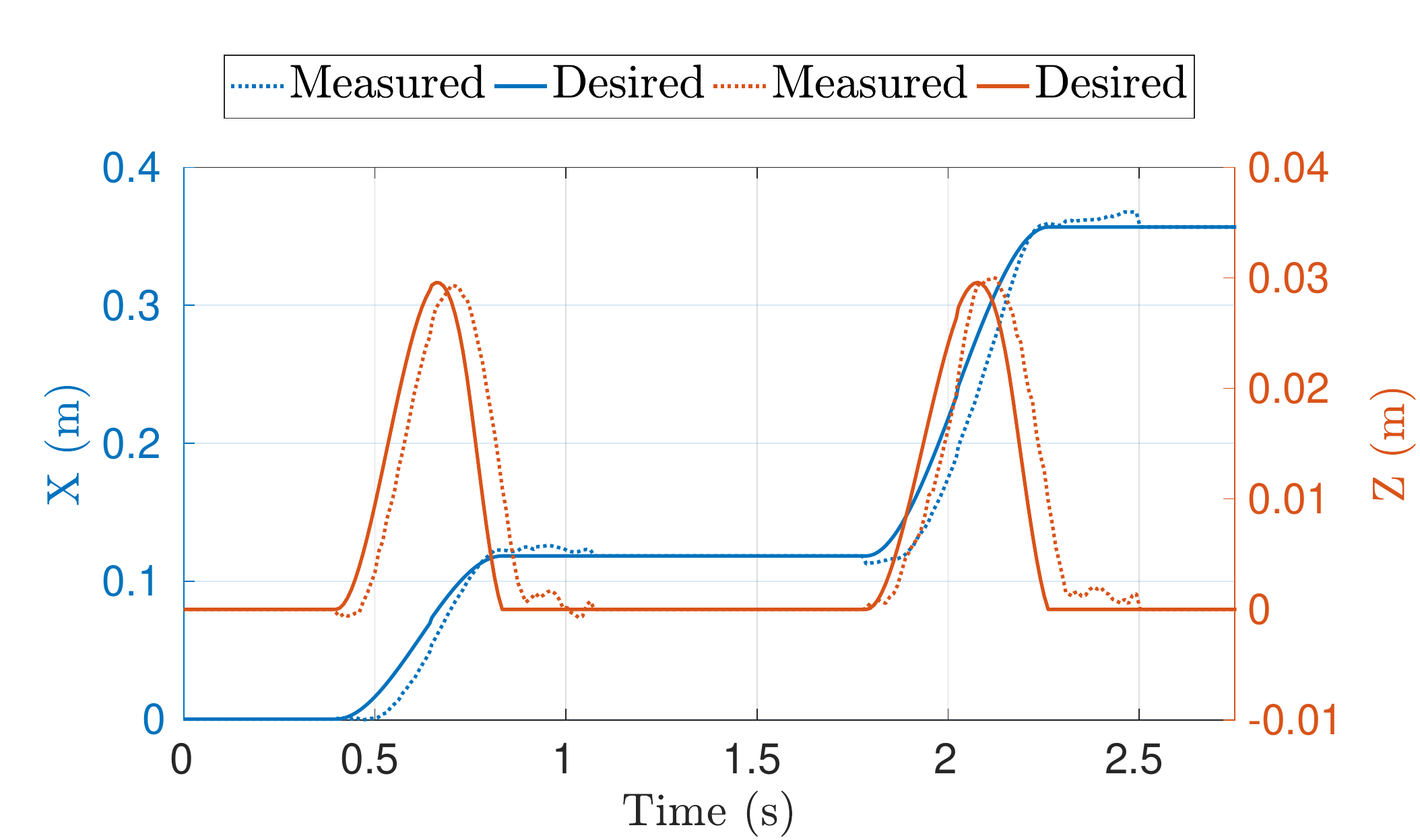}
            \caption{ }
            \label{fig:inst_pos-max_vel-lf}
            \end{subfigure}
        \end{myframe}
    \end{subfigure}
    \begin{subfigure}[b]{\columnwidth}
    \begin{myframe}{Instantaneous + Velocity Control}
        \begin{subfigure}[t]{0.49\columnwidth}
        \centering
        \includegraphics[width=\textwidth]{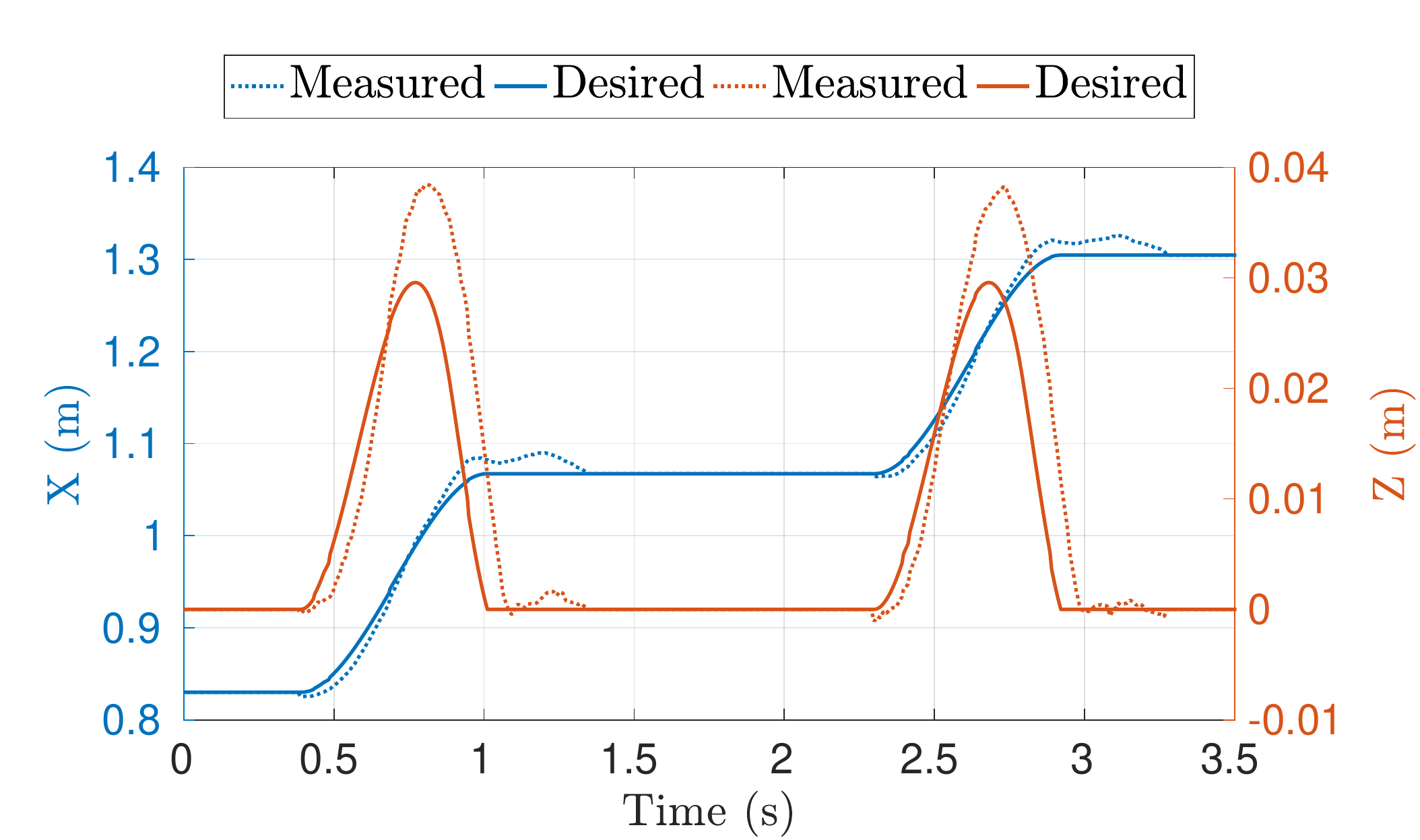}
        \caption{ }
        \label{fig:inst_vel-min_vel-lf}
    \end{subfigure}
    \begin{subfigure}[t]{0.49\columnwidth}
        \centering
        \includegraphics[width=\textwidth]{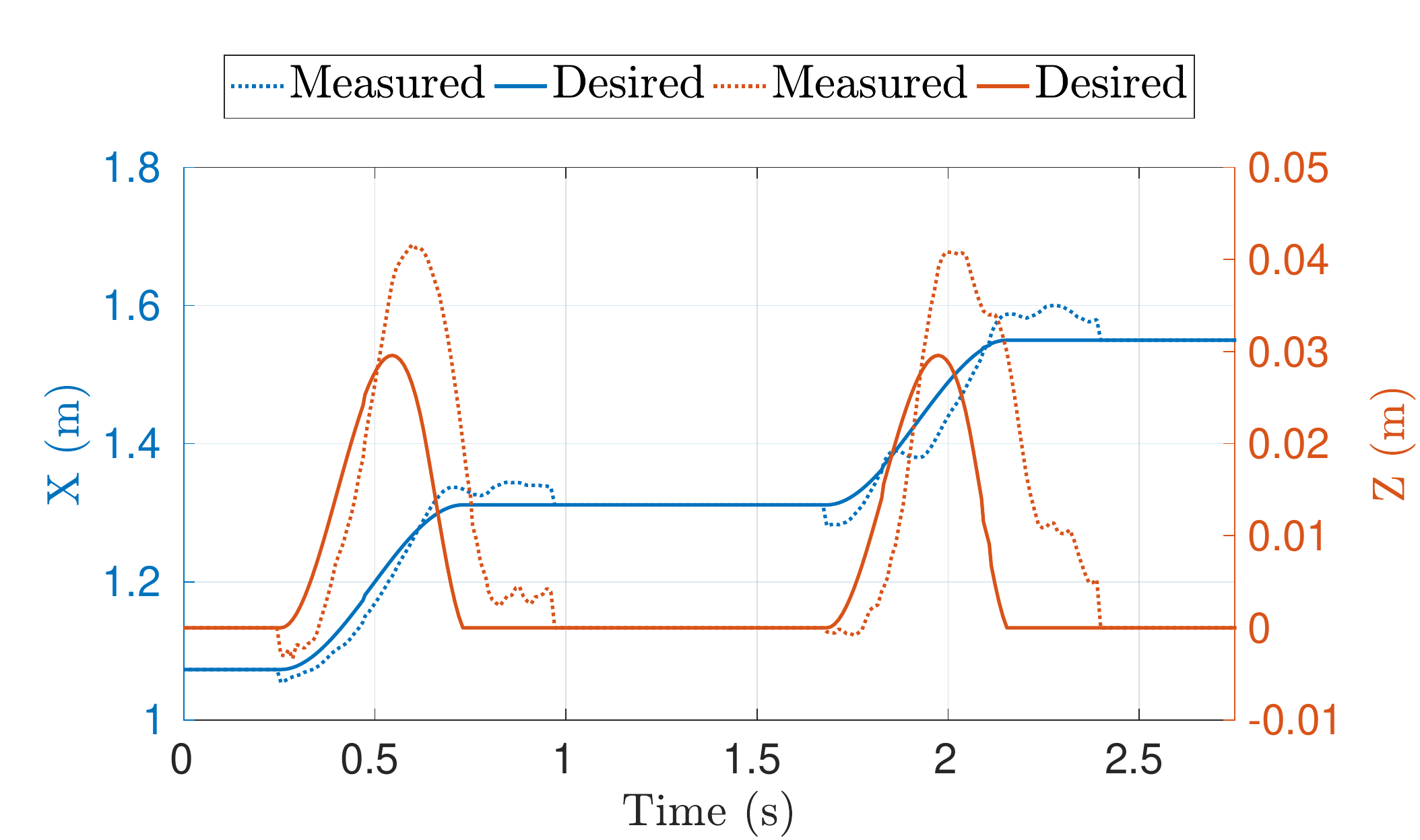}
        \caption{ }
        \label{fig:inst_vel-max_vel-lf}
    \end{subfigure}
    \end{myframe}
        \end{subfigure}
     \caption{Tracking of the left foot position using the instantaneous simplified model control. Whole-body QP  control as position control (a) and  velocity control (c),  walking velocity:  $\SI{0.19}{\meter \per \second}$. Whole-body QP control as position control (b) and  velocity control (d), walking velocity: $\SI{0.41}{\meter \per \second}$. }
     \vskip-0.25cm
\end{figure*}

\subsection{Simplified model control: Predictive versus Instantaneous}
\label{sub_sec:Simplified Models Controller}

In this section, we compare the control laws~\eqref{eq:reactive_dcm} and~\eqref{MPC_solution}, which both generate a (desired) center-of-pressure that attempt at stabilizing  a desired DCM. To simplify the comparison, the controller of the \emph{whole-body QP layer} is kept fixed in this section, and we show and discuss only the results when the robot is position controlled.
The time horizon of the predictive control is $\SI{2}{\second}$.
\subsubsection{Experiment 1}

Figs. \ref{fig:inst_pos-min_vel-dcm} and \ref{fig:mpc_pos-min_vel-dcm} depict the DCM tracking performances obtained with the instantaneous and predictive controllers, respectively. Both controllers seem to show good tracking performances, and the DCM error is kept below $\SI{5}{\centi \meter}$ in both cases. Note that the instantaneous controller induces, however, faster variations of the measured DCM. This contributes to overall higher vibrations of the robot. One of the reasons for this variation is that the instantaneous controller~\eqref{eq:reactive_dcm} injects a (desired) center-of-pressure proportional to the measured DCM $\xi = x + \dot{x}/\omega$, which in turns contains the center-of-mass velocity. To mitigate this, we  suggest to filter joint velocities appropriately. In our case, however, joint velocities were not filtered to avoid delays in the measured DCM. Our experience showed that adding a filter on joint velocities is not an easy task, and we did not find the right trade off for obtaining overall performance improvements. 

Figs.~\ref{fig:inst_pos-min_vel-com} and \ref{fig:mpc_pos-min_vel-com}, depict the CoM tracking performances, which are mainly dependent on the ZMP-CoM controller~\eqref{eq:zmp_controller}. This controller receives desired DCM values  from the \emph{simplified model control} layer, which are obtained with either the instantaneous or predictive controllers. In both cases, CoM tracking performances are good, and the CoM error is kept below $\SI{2}{\centi \meter}$.

Figs.~\ref{fig:inst_pos-min_vel-zmp} and~\ref{fig:mpc_pos-min_vel-zmp} depict the ZMP tracking performances, which are still mainly dependent on the ZMP-CoM controller~\eqref{eq:zmp_controller}. It is important to observe that the desired ZMP is smoother when the \emph{simplified model control} uses the predictive law~\eqref{MPC_solution} to generate it. This is a tunable property that depends on the associated weight in the cost function of the MPC problem. Although this smoother behaviour does contribute to less robot vibrations, the overall robot performance became less reactive and, consequently, less robust to robot falls. Although the extensive hand-made tuning, we were not able to increase the robot velocity when the \emph{simplified model control} used the predictive law~\eqref{MPC_solution}. 

\subsubsection{Experiment 2}

At a robot desired walking speed of $\SI{0.41}{\meter \per \second}$, there is initially no significant difference between the DCM tracking obtained with instantaneous and predictive control laws -- see Fig.~\ref{fig:mpc_pos-max_vel-dcm} and~\ref{fig:inst_pos-max_vel-dcm} for $t < \SI{1.5}{\second}$. However, fast robot walking velocities require fast variations of the desired CoM and ZMP. In the case of the predictive controller, this fast variation induces a not-very-good performance of the desired ZMP around $t = \SI{1.5}{\second}$ -- see Fig.~\ref{fig:mpc_pos-max_vel-zmp}. Clearly, these bad performances in turn induce a bad tracking of the DCM shown in Fig.~\ref{fig:mpc_pos-max_vel-dcm} at $t\approx \SI{2}{\second}$, and consequently a robot fall. At this point, one is tempted to increase the gain $K_{zmp}$ of the controller~\eqref{eq:zmp_controller}, which shall induce a better tracking of the ZMP. This unfortunately gives raise to higher robot oscillations, which in turn degrades the ZMP-CoM tracking, and still a robot fall. 

In a nutshell, the \emph{predictive simplified control} is much less robust than the  \emph{instantaneous simplified control} with respect to ZMP tracking errors. We suggest to filter the measurements from force sensors to obtain less noisy ZMP measurement, which should allow one to increase the gains for ZMP tracking. In our case, adding filters led to slower system response and, consequently, a robot fall.

\subsection{Whole-Body QP Control: Position versus Velocity}
In this section, we compare the performances of the three layer architecture when the robot is either position or velocity controlled -- see Sec.~\ref{subsubsec-pos-vel-control} for the meaning of these  modes. For comparison purposes, the \emph{simplified model control} is achieved via the instantaneous control~\eqref{eq:reactive_dcm}.  Also, to emphasize the comparison, we stress the importance of the tracking of the desired feet positions. 

\subsubsection{Experiment 1}
Figs. \ref{fig:inst_pos-min_vel-lf} and \ref{fig:inst_vel-min_vel-lf} depict the tracking of desired left foot positions when the robot is either position or velocity controlled, respectively. The position controller ensures better tracking performance than the velocity one. One is then tempted to increase the gains of the * quantities~\eqref{feetVelocitiesStar}, which should increase the tracking performances of the velocity control. However, we observed that the noise due to numerical derivative is harmful for the overall performance, and makes the robot shake and then fall. Again, filtering the taken joint measures may be helpful, but it introduced a delay 
that did not allow us to increase the walking speed.

\subsubsection{Experiment 2}
The aforementioned foot position tracking problem worsens at higher walking velocity. Fig.~\ref{fig:inst_pos-max_vel-lf} shows  that the feet tracking error is lower than $\SI{5}{\centi \meter}$ on the $x$ axis and $\SI{1}{\centi \meter}$ on the $z$ one for position control. Instead, the velocity control in Fig.~\ref{fig:inst_vel-max_vel-lf} keeps the error always lower than $\SI{6}{\centi \meter}$ on the $x$ component and $\SI{3}{\centi \meter}$ on the $z$ direction. 

%% file: tex/conclusion.tex
\section{CONCLUSION AND FUTURE WORK}
\label{sec:CONCLUSION_AND_FUTURE_WORK}
This paper contributes towards the benchmarking of different implementations of state-of-the-art control architectures for humanoid robots locomotion. In particular, the control architecture is composed of three layers, which all exploit the concept of the Divergent Component of Motion. 
The three layers  are here called: trajectory optimization,  simplified model control, the whole-body QP control.
A key feature of this paper is that we compare walking results obtained with predictive and instantaneous controllers for the simplified model layer. Also, we compare position and velocity robot control modes for the whole-body QP control layer.
We show that instantaneous controllers combined with robot position control allowed us to achieve a desired walking speed of $\SI{0.41}{\meter \per \second}$, which is the highest walking velocity ever achieved for the iCub humanoid robot.

As future work, we plan to extend the proposed benchmarking to torque-control algorithms (e.g.~\cite{7803266}), and to propose architecture implementations allowing the robot walking on inclined planes. 
Another interesting future work is the implementation of a dynamic footstep planner.
Indeed, currently, the footsteps are generated independently from the state of the robot.
A dynamic planner can plan the footprints according to the external disturbance acting on the humanoid, i.e. an unknown contact force acting on the robot.
